\documentclass[11pt]{article}

\usepackage{microtype} 
\usepackage{booktabs}  
\usepackage{url}  
\usepackage{graphicx}

\usepackage{amsmath}
\usepackage{amsthm}

\usepackage{algorithm}
\usepackage{algpseudocode}

\usepackage[final, workshop]{automl}

\usepackage{natbib}
\title{Investigating the Impact of Hard Samples on Accuracy Reveals In-class Data Imbalance}

\author[1]{\nameemail{Pawel Pukowski}{ppukowski1@sheffield.ac.uk}}
\author[1]{\nameemail{Haiping Lu}{email1@example.com}}

\affil[1]{University of Sheffield}

\hypersetup{%
  pdfauthor={}, 
  pdftitle={},
  pdfsubject={},
  pdfkeywords={}
}

\begin{document}

\maketitle

\begin{abstract}

In the AutoML domain, test accuracy is heralded as the quintessential metric for evaluating model efficacy, underpinning a wide array of applications from neural architecture search to hyperparameter optimization. However, the reliability of test accuracy as the primary performance metric has been called into question, notably through research highlighting how label noise can obscure the true ranking of state-of-the-art models. We venture beyond, along another perspective where the existence of hard samples within datasets casts further doubt on the generalization capabilities inferred from test accuracy alone. Our investigation reveals that the distribution of hard samples between training and test sets affects the difficulty levels of those sets, thereby influencing the perceived generalization capability of models. We unveil two distinct generalization pathways—toward easy and hard samples—highlighting the complexity of achieving balanced model evaluation. Finally, we propose a benchmarking procedure for comparing hard sample identification methods, facilitating the advancement of more nuanced approaches in this area. Our primary goal is not to propose a definitive solution but to highlight the limitations of relying primarily on test accuracy as an evaluation metric, even when working with balanced datasets, by introducing the \textit{in-class data imbalance problem}. By doing so, we aim to stimulate a critical discussion within the research community and open new avenues for research that consider a broader spectrum of model evaluation criteria. The anonymous code is available at \url{https://github.com/PawPuk/CurvBIM} blueunder the GPL-3.0 license.
\end{abstract}

\section{Introduction}

\begin{figure}[t]
    \begin{subfigure}{3.5cm}
        \centering
        \includegraphics[width=3.5cm]{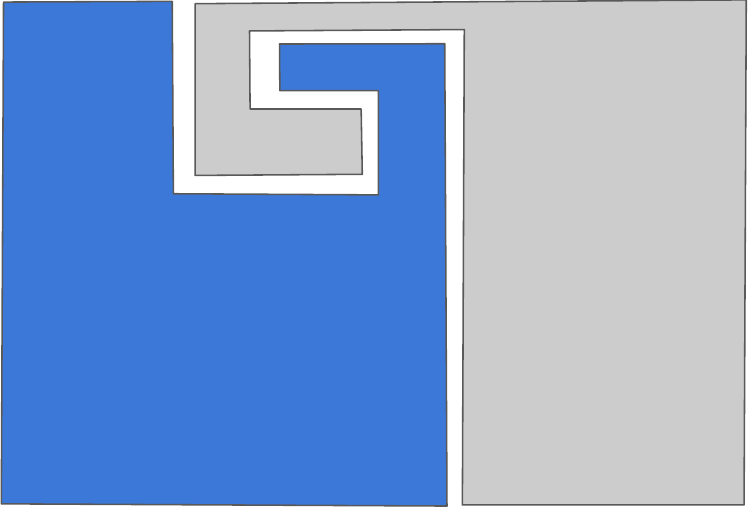}
        \caption{Theoretical scenario}
    \end{subfigure}
    \begin{subfigure}{3.5cm}
        \centering
        \includegraphics[width=3.5cm]{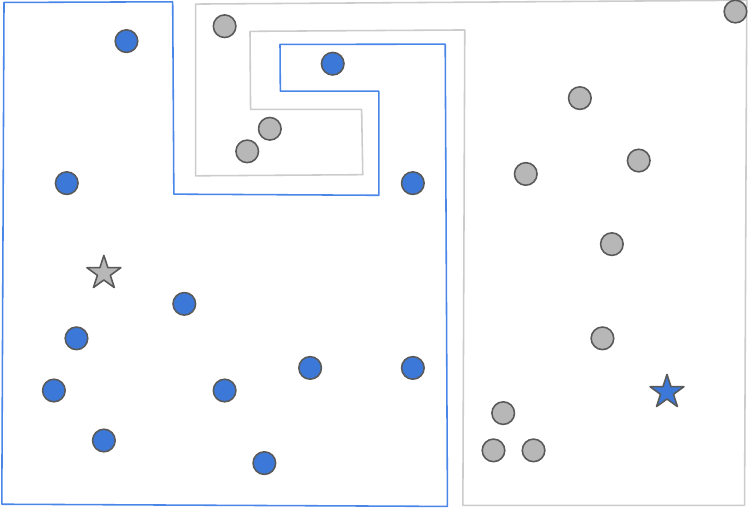}
        \caption{Real-world scenario}
    \end{subfigure}
    \begin{subfigure}{3.5cm}
        \centering
        \includegraphics[width=3.5cm]{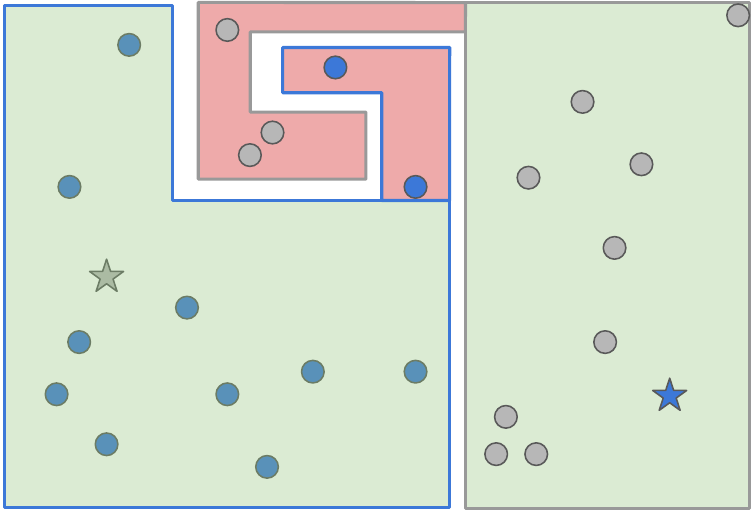}
        \caption{Easy-hard regions}
        \label{fig:fig1c}
    \end{subfigure}
    \begin{subfigure}{4.5cm}
        \centering
        \includegraphics[width=4.5cm]{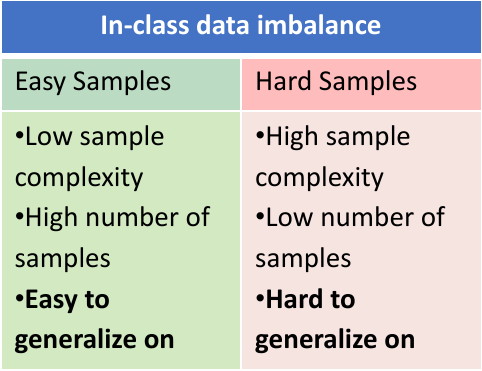}
        \caption{In-class data imbalance}
    \end{subfigure}
    \caption{Let's consider a hypothetical binary classification scenario (a) featuring two distinct class manifolds. The real-world equivalent of this scenario would involve a point cloud non-uniformly sampled from the class manifolds, with some added label noise (stars in b). In this work, we propose that the difficulty of training is derived from the geometrical and topological properties of the class manifolds, leading to areas with higher/lower sample complexity (red/green respectively), due to factors like curvature and homology. Consequently, we observe the emergence of the in-class data imbalance problem (d), which stems from the fact that, although it is more challenging to learn from hard samples because of their sample complexity, the datasets are predominantly composed of easy, not hard, samples.}
    \label{fig:fig1}
\end{figure}

The pivotal role of test set performance in machine learning is undeniable. In AutoML the accuracy on test sets guides the automated selection \citep{lim2019fast, kerssies2023neural}, tuning \citep{makarova2022automatic}, and evaluation of models \citep{ericsson2023better}, ensuring that the systems developed are both robust and capable of adapting to new, unseen challenges. Yet, this evaluation method is not immune to the challenges posed by data problems, such as dataset imbalance and label noise. Traditional concerns of imbalance refer to disproportionate representation of classes within the data, where one class significantly outnumbers others, which can cause models to perform well on majority classes while neglecting minority ones, misleadingly inflating perceived generalization capabilities \citep{li2021autobalance, chawla2002smote}. Label noise, as highlighted by \citet{northcutt2021pervasive}, introduces additional uncertainty, corrupting the test data with incorrect labels, which can significantly affect the perceived model performance (see Figure \ref{fig:fig1}a). In this paper, we shed light on an overlooked aspect that further complicates the reliability of test set performance: \textit{the distribution of hard and easy samples across training and test sets}.

The manifold hypothesis, which states that high-dimensional data tend to occupy lower-dimensional manifolds \citep{fefferman2016testing}, provides a framework for distinguishing between easy and hard samples in machine learning contexts. In this context, inference involves disentangling class manifolds \citep{naitzat2020topology}, and learning aims at finding the appropriate space transformations for this task. Consequently properties such as intrinsic dimension \citep{ansuini2019intrinsic, pope2021intrinsic}, persistent homology \citep{birdal2021intrinsic}, curvature \citep{kienitz2022effect} and entanglement \citep{DBLP:conf/icml/KaufmanA23} of these manifolds serve as complexity measures. The fact that those characteristics are often local rather than global, implies that the sample complexity differs across each class manifold, resulting in areas that are easier and harder to learn (see Figure \ref{fig:fig1c}). Throughout this work, we conceptualize class manifolds as topological structures populated by samples of a specific class, encompassing both observed samples in the dataset and hypothetical, unobserved ones. Furthermore, drawing inspiration from \citet{narayanan2009sample}, we define sample complexity as the number of random samples from a given neighborhood within a class manifold needed to reliably classify a new data sample from the same neighborhood into one of the predefined classes with high probability.

Building on this geometric foundation, our work reveals an intriguing observation: \textit{the localised property of sample complexity paired with non-uniform sampling prevalent in real-world scenarios results in an in-class data imbalance within individual class manifolds} (see Figure \ref{fig:fig1c}). In our examination of in-class data imbalance, we observe foundational similarities to classical, between-class data imbalance problems, characterized by the existence of two distinct groups within a dataset: one exhibiting high accuracy on the test set and the other showing poor accuracy. Additionally, we find that there's a notable discrepancy in sample quantities, with the easy group vastly outnumbering the hard group by as much as 13 times on MNIST \citep{deng2012mnist} and 6 times on FashionMNIST \citep{xiao2017fashion} and KMNIST \citep{clanuwat2018deep}. 

Delving deeper, we find that, unlike in the between-class data imbalance problem, the differences in performance are not solely a matter of sample quantity but also reflect the differing complexities between these groups. This is apparent in our experiments: adding 3,250 easy samples boosted accuracy on easy samples in the test set from 47\% to 80\%, whereas incorporating 5,000 hard samples improved the accuracy on hard samples in the test set from 17\% to only 24\%, when measured on MNIST. Another distinction between in-class data imbalance and the between-class data imbalance problem arises from the fact that both hard and easy samples originate from the same class, thereby facilitating some level of information sharing between them. Yet, this sharing is uneven and not dataset-consistent, as evidenced by our empirical results: training exclusively on hard samples led to 50\% accuracy on easy samples in the test set, whereas focusing solely on easy samples during training resulted in a considerably lower accuracy on hard samples in the test set, at 15\% on MNIST. Furthermore, increasing the ratio of hard samples in the training set has a positive effect on the accuracy of easy samples from the test set on MNIST and KMNIST, \textit{but not on FashionMNIST}, suggesting dataset-dependent level of information sharing.

The existence of in-class data imbalance indicates the need for a more thoughtful strategy in our data practices, from augmentation to dataset compilation and sample division into training and test sets, challenging the status quo of random splitting.

In this regard, our contributions are three-fold:
\begin{itemize}
    \item \textbf{Impact of In-class Data Imbalance}: Through empirical analysis, we demonstrate the existence of two distinct groups within the class manifold, highlighting how the distribution of hard samples between training and test sets impacts model training difficulty and the perceived generalization performance.
    \item \textbf{Identifying Hard Samples}: We adapt the stragglers concept from \citet{ciceri2024inversion}—samples misclassified at the \textit{inversion point}, where class manifolds transition from expanding to contracting, indicating a shift from learning easy to harder samples—to the multiclass classification setting. This extension provides a nuanced approach to identifying and learning from hard samples, enriching our understanding of sample complexity within diverse classification contexts.
    \item \textbf{Benchmarking Procedure for Hard/Easy Sample Segregation}: We introduce a benchmarking procedure for evaluating methods that segregate hard from easy samples, facilitating a standardized assessment of strategies aimed at addressing in-class data imbalance.
\end{itemize}

\section{Relevant Literature}
\label{sec:relevant_literature}

In machine learning, hardness is contextually defined. In SVMs, hardness is attributed to samples that lie close to the decision boundary, critical for defining the optimal separating hyperplane. As such they utilize geometric margins as a direct measure of sample difficulty \citep{cortes1995support}. Active learning defines hard samples as those that offer a balance between high uncertainty and representativeness, maximizing learning efficiency and distribution coverage \citep{budd2021survey, ren2021survey}. 

Curriculum learning and active learning both often rely on uncertainty-based metric to asses the hardness, but due to the different usage of the hardness notion their techniques developed in different directions. Curriculum learning, first introduced by \citet{bengio2009curriculum}, employs a structured approach, progressively increasing task difficulty based on the model's current confidence, often utilizing competence-based metrics not typically found in active learning \citep{wang2021survey, soviany2022curriculum}. Active learning, in contrast, selects samples based on expected model change, a metric unique to its methodology, aiming to label those that most improve the model \citep{settles2009active, zhou2023samples}. In hard negative mining, hardness is defined by samples that are incorrectly classified as positive, focusing on misclassification rate and loss values to identify the most challenging examples \citep{canevet2015efficient, shrivastava2016training}. This emphasizes the selection of negatives that contribute most to learning, diverging from the uncertainty-based metrics commonly used in other domains.

In the literature, adversarial examples and anomalies are frequently identified as hard samples \citep{ho2020contrastive, yang2023dcdetector}. This classification stems from the inherent challenges these samples present in model training and generalization. Adversarial examples, intentionally engineered to exploit model weaknesses, underscore the limitations in current learning algorithms' ability to withstand malicious inputs \citep{goodfellow2014explaining, brendel2017decision, serban2020adversarial}. Conversely, anomalies represent outliers within the data distribution, which are naturally occurring yet difficult for models to accurately classify due to their rarity or deviation from the norm \citep{hodge2004survey, chandola2009anomaly, goldstein2016comparative}.

Our work is inspired by \citet{ciceri2024inversion}, who introduced the notion of \textit{stragglers} in binary classification contexts. They defined stragglers as points misclassified by a model at a specific phase—when it shifts from decreasing intra-class variability (while increasing inter-class variability) to doing the opposite. Their focus was on establishing the presence of stragglers and demonstrating their effect on generalization. We extend their methodology to multiclass classification settings (Section \ref{sec:generalization_of_stragglers}). However, our work diverges significantly from \citet{ciceri2024inversion} beyond this point, as we focus on studying the in-class imbalance (Section \ref{sec:strategic_sample_distribution}), and propose a benchmarking procedure for hard sample identifiers (Section \ref{sec:benchmark}).

Our work also bears a close resemblance to research on small disjuncts \citep{holte1989concept, japkowicz2001concept, jo2004class}. These studies suggest that the problem of between-class imbalance is not an issue in itself, but rather that it is accompanied by the problem of small disjuncts. Each disjunct can be considered a submanifold of the class manifold, with small disjuncts corresponding to submanifolds (or neighborhoods) that are sparsely sampled. The sparse sampling often means they are harder to learn, although this depends on their sample complexity. We argue that there is significant merit in rekindling interest in this topic through the lens of sample complexity introduced in our paper. This is because the majority of works on within-class imbalance address the imbalance issues using density-related variables \citep{anwar2014measurement, li2024multi}. We contend that this approach oversimplifies the problem, and that sample complexity offers a more thorough and robust perspective. We hope this insight will generate increased interest in this under-researched area \citep{leevy2018survey}.

\section{Towards Objective Identifiers of Hard Samples}
\label{sec:generalization_of_stragglers}

Reflecting on the diverse interpretations of hardness examined in Sec. \ref{sec:relevant_literature}, it becomes clear that most definitions are tailored to specific contexts within machine learning. This observation underscores an absence of a hardness measure that is objective, rooted in the data itself, and transferable across different learning contexts. To bridge this gap, we extend and refine the methodology introduced by \citet{ciceri2024inversion}, shifting our focus towards a detailed examination of hard regions within datasets that inherently challenge learning algorithms.

\noindent\textbf{Background:} \citet{ciceri2024inversion} document non-monotonous development of the radii of class manifolds in binary classification setting see Appendix \ref{sec:radii_description_appendix} for description of the radii metric). Initially, these manifolds diverge, enhancing intra-class variability while concurrently reducing their radii, effectively minimizing inter-class variability. This phase transitions at a critical juncture that the authors call the \textit{inversion point}, beyond which the trend reverses. This inversion point serves as a marker for identifying \textit{stragglers}, a unique subset of data samples characterized by their misclassification by the model at this pivotal moment. Crucially, \citet{ciceri2024inversion} also demonstrate that stragglers are nearly universal, showing a substantial overlap between sets identified under different models or optimisation settings. In conjunction with the research of \citet{arpit2017closer}, which identifies distinct learning paces for easy and hard examples in real datasets, these results support a conjecture of a two-stage learning process: initially concentrating on easy samples before transitioning to a focus on the harder ones.

\noindent\textbf{Multiclass scenario:} While \citet{ciceri2024inversion} laid the groundwork with their analysis, it is crucial to acknowledge that they focused on binary classification. This raises an intriguing question about the applicability and generalization of their findings to the more complex scenario of multiclass classification tasks. To address this gap, we generalize the methodology established by \citet{ciceri2024inversion} to datasets in their original, multiclass setting. We focus on the radii metric, as calculating it is less computationally intensive than calculating inter-class distances in multiclass setting. We trained an ensemble of $100$ Fully Connected Neural Networks (FC NN), using the same experimental set-up as \citet{ciceri2024inversion}, with the critical distinction of utilizing the entire dataset for both training and testing. This approach was deliberately chosen to ensure comprehensive identification of stragglers across the whole dataset rather than a subset. While this method inherently increases the likelihood of overfitting, we maintain that this aspect does not detract from our findings. We posit that the difficulty attributed to hard samples is linked to their data-derived characteristics, a factor that remains unaffected by the degree of fit between the model and the data.In Appendix \ref{sec:Replicating_inversion_point_appendix} we present a pseudocode and experimental setup.

\begin{figure}[t]
    \includegraphics[width=15cm]{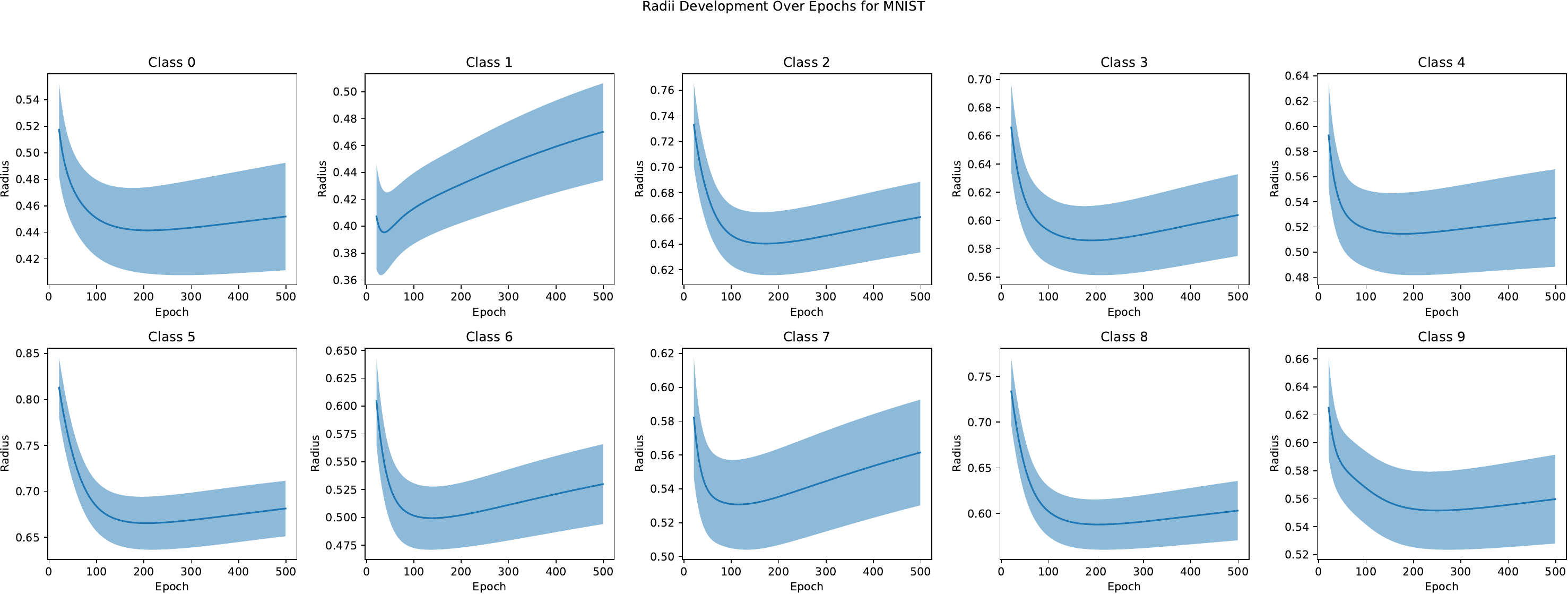}
    \caption{By generalizing the method introduced by \citet{ciceri2024inversion} to multiclass MNIST, distinct inversion points emerge for each class. This observation signifies that the dynamics of manifolds segmentation are class-specific—a nuance not captured by before due to previous focus on the binary classification setting. The results on other datasets are available in Appendix \ref{sec:generalization_of_stragglers_appendix}.}
    \label{fig:fig2}
\end{figure}

\noindent\textbf{Analysis:} A detailed examination of our results (Figure \ref{fig:fig2}) reveals patterns that corroborate the presence of an inversion point across all class manifolds during training. We find a global minimum for all classes, confirming that the inversion point is a recurring feature in the multiclass classification context. However, this inversion point is not uniform across classes. For instance, in the MNIST dataset, the class manifold for digit $1$ reaches this inflection point sooner than others, indicating a class-dependent onset of phase two in the training. We hypothesize that this happens due to the distinct geometrical and topological properties of class manifolds, such as entanglement and intrinsic dimension.

We also notice that the contrast between the radii at the start of training and at the inversion point is markedly greater than what was reported in the binary classification scenarios, and also seems to be class-dependent. This suggests that the process of intra-class contraction and inter-class divergence is more pronounced when dealing with a full spectrum of class labels.

Lastly, while most classes exhibit an inversion point, it is noteworthy that in some instances an inversion point is not evident. Although such cases are rare, they present an important caveat to the universality of the inversion phenomenon and hint at the potential for certain classes to follow alternative learning dynamics that do not conform to the expected pattern. These findings highlight the complexity of learning dynamics at play and reinforces the notion that a \textit{one-size-fits-all approach may not be sufficient when dealing with multiclass datasets} \citep{cheung2021adaaug, balestriero2022effects}.

\section{In-class Data Imbalance}
\label{sec:strategic_sample_distribution}

\begin{figure}[t]
    \centering
    \includegraphics[width=15cm]{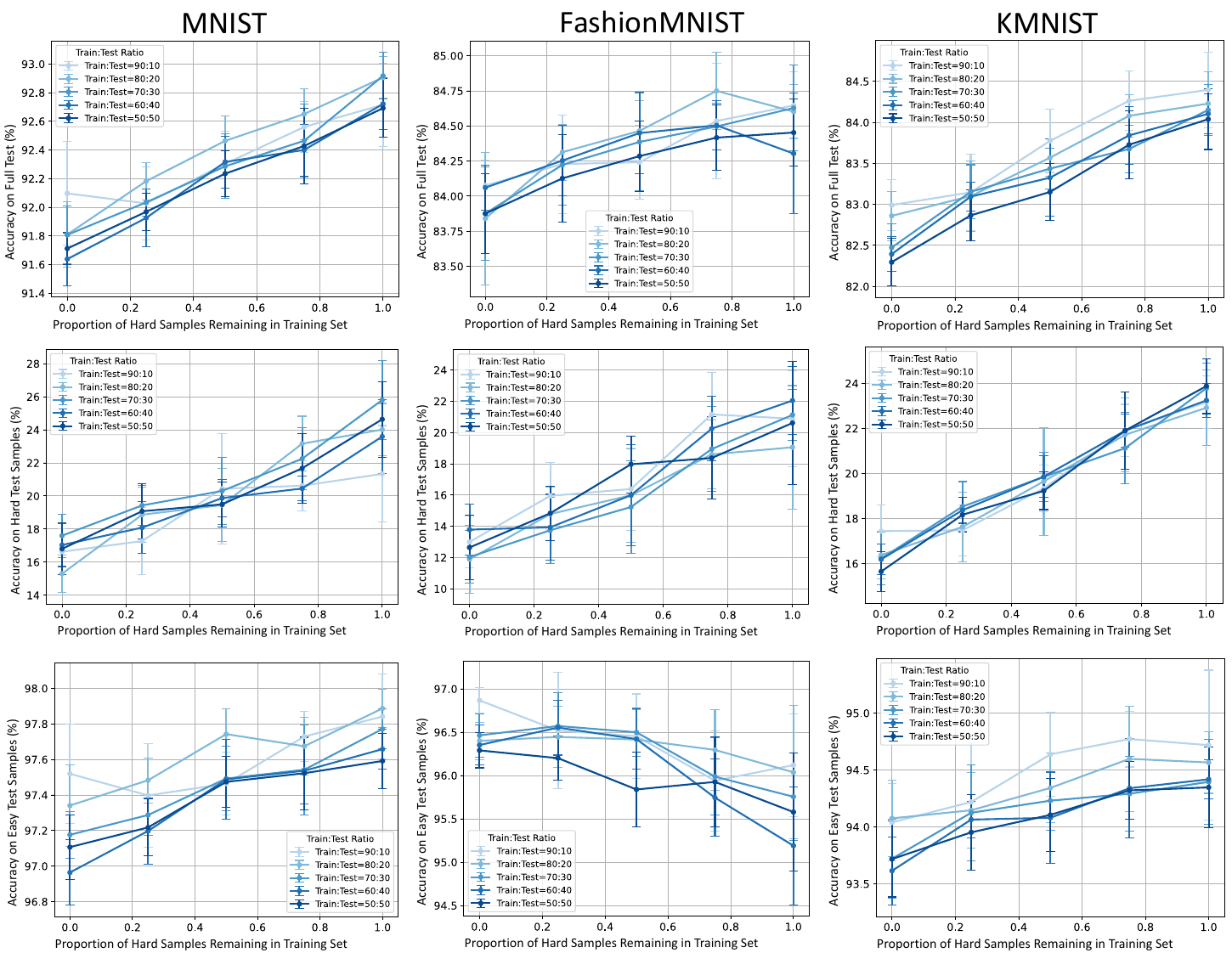}
    \caption{Increasing the proportion of hard samples in the training set improves accuracy on the entire test set (Row 1), but the difference comes mostly from improved accuracy on other hard samples (Row 2), rather than easy samples (Row 3).}
    \label{fig:fig3}
\end{figure}

\begin{figure}[t]
    \centering
    \includegraphics[width=15cm]{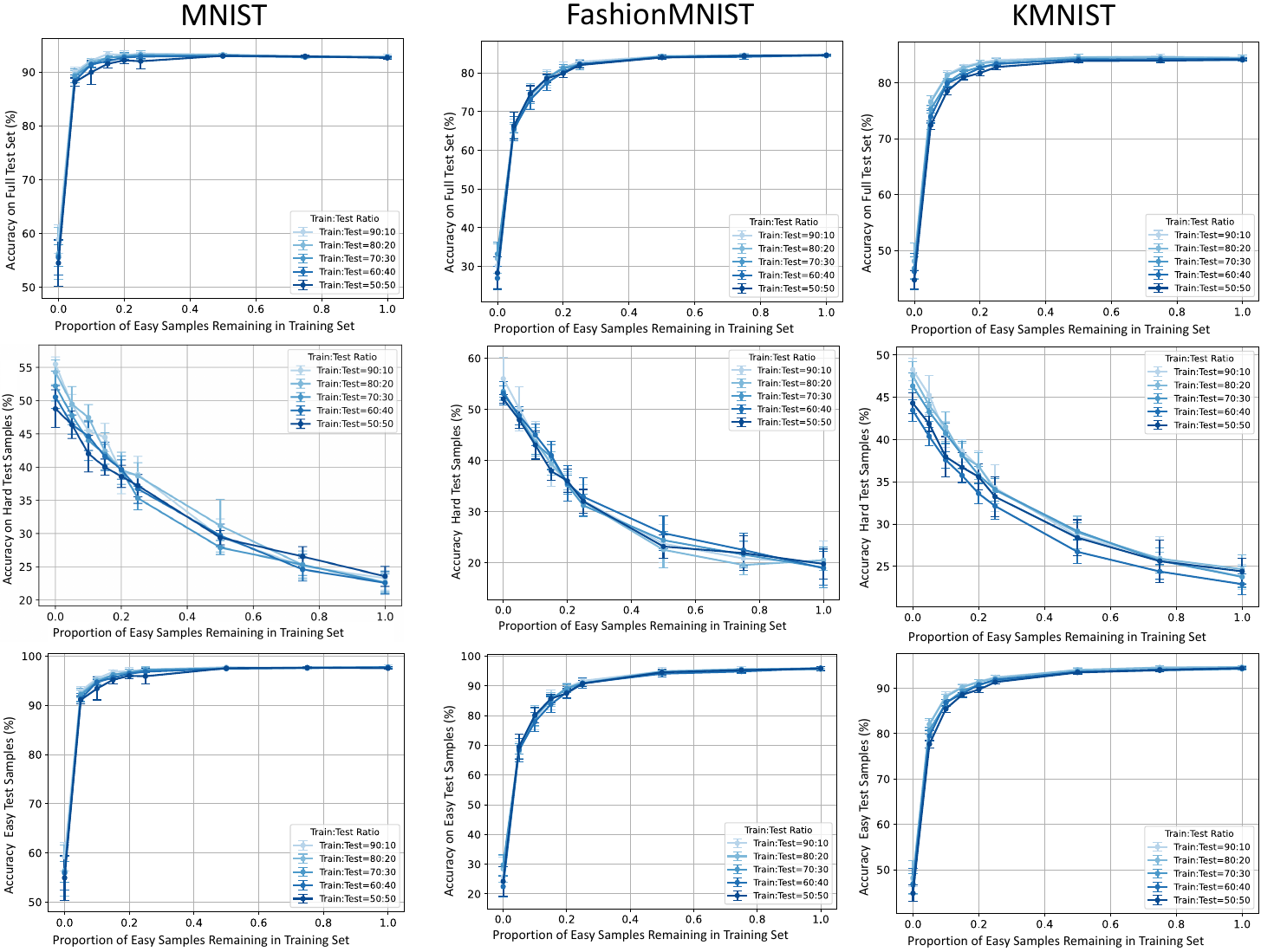}
    \caption{Increasing the proportion of easy samples in the training set improves accuracy on easy samples in the test set (Row 3), resulting in increased overall accuracy (Row 1), while decreasing the accuracy on hard samples in the test set (Row 2), which resembles results we would get when adding samples from the majority class in between-class data imbalance problem.}
    \label{fig:fig4}
\end{figure}

\textbf{Experimental setup:} In this section we use stragglers to identify hard samples, an intuitive method now empirically proven to effectively mark complex cases. To obtain a straggler set, we train a model until $k$ inversion points are found, one for each class. These result in $k$ class-level straggler sets, which we later aggregate to form a dataset-level straggler set. We then distribute stragglers across training and test sets in proportions reflecting train-to-test ratios. For each ratio, we create three unique sets of stragglers. We then train three models with distinct initializations for each straggler set. To explore the impact of the quantity of hard and easy samples on the perceived generalization of the models, we conduct the experiments multiple times, each time removing a certain percentage of hard (see Fig. \ref{fig:fig3}) or easy (see Fig. \ref{fig:fig4}) samples from the training set. In other words we start with the full dataset, and gradually remove hard (see Fig. \ref{fig:fig3}) or easy (see Fig. \ref{fig:fig4}) samples from the training set. In Appendix \ref{sec:verifying_importance_of_stragglers_appendix} we present a pseudocode and experimental setup.

\noindent\textbf{Similarities to between-class data imbalance:} The observed ratios of stragglers to non-stragglers in our datasets—1:6 for FashionMNIST and KMNIST, and a more extreme 1:13 for MNIST—reflect an in-class data imbalance reminiscent of the classical majority:minority dynamic. This similarity extends to the performance on the test set, with hard samples consistently showing lower accuracy (18-28\% on MNIST) compared to easy samples (97-98\% on MNIST), akin to the disparity seen between majority and minority classes. Moreover, the way to counteract this imbalance appears similar to class imbalance strategies: introducing more samples from the specific group one wishes to improve accuracy on—be it hard or easy—yields the most significant gains. This suggests that \textit{in scenarios where data augmentation is employed, prioritizing the inclusion of hard samples might yield more substantial improvements}. This insight opens avenues for future exploration, particularly in integrating such considerations into Automatic Data Augmentation \citep{cubuk2018autoaugment, lim2019fast}.

\noindent\textbf{Nuances emerging from in-class data imbalance:} The similarities between in-class and between-class data imbalance problems lead to an inevitable question: Are the approaches commonly used to address between-class data imbalance applicable to the in-class data imbalance problem? As an example, let's consider the F1 score, which is arguably the most popular metric used when working with multiclass imbalanced datasets. The F1 score primarily evaluates the balance between precision and recall, effectively addressing between-class imbalances by penalizing models that overly favor one class over the others. However, this metric assumes uniform difficulty within classes, a presumption that overlooks the nuanced reality of in-class data imbalances where samples vary significantly in complexity. The F1 score, while indicating overall balance in datasets like MNIST, fails to capture these subtleties. This observation is supported by our results, which we detail in Appendix \ref{sec:f1_investigation_appendix}.

The initial distinction between in-class and between-class data imbalance arises from the fact that both hard and easy samples originate from the same class, inherently sharing a level of information. However, our experiments show that this information is unevenly shared. We notice that training solely on hard samples considerably boosts accuracy on easy samples (50-61\% on MNIST). Conversely, training with only easy samples yields markedly lower accuracy on hard samples (14-19\% on MNIST). These results become even more surprising when we consider that hard samples constitute a minority within the dataset.

The second significant observation from our experiments underscores the role of sample complexity in learning performance. Remarkably, a modest pool of merely 3,000 easy samples is sufficient to elevate the accuracy on easy samples, from 50-61\% to 90-93\% on MNIST. This enhancement starkly contrasts with the marginal gains observed when incorporating all 5,000 hard samples on MNIST (improvement from 14-19\% to 18-28\%). The impact of sample complexity becomes even more apparent when we study the curves in Figure \ref{fig:fig3}c, with the function being the most smooth for FashionMNIST, which is the most complex of the tested datasets.

We also find that the impact of adding hard samples on the accuracy on easy samples varies across datasets, suggesting that the level of information that hard samples have on easy samples might differ from dataset to dataset. We see that while MNIST and KMNIST benefit from the inclusion of hard samples, \textit{FashionMNIST exhibits a decline in the performance}, possibly due to its higher complexity or more prevalent label errors. Interestingly, the train:test ratio has no impact on our results. This suggests that hard and easy samples exhibit distinct generalization capabilities, with a certain amount of shared information between these groups that appears to be dataset-dependent.

Finally, we investigated the distribution of hard samples across classes, our findings (Figure \ref{fig:final_figure}) reveal a non-uniform distribution strongly correlated with class-level accuracies. In fact, the Pearson correlation coefficient between the class-level error rates and the class-level distribution of hard samples is above 0.9 for all hard sample identifiers and datasets used in our paper (see Appendix \ref{sec:f1_investigation_appendix} for more information). Furthermore, variations in observed in-class data imbalance, based on the choice of hard sample identifier, are attributed to the significant overlap among hard samples. Specifically, there is an $87.42\% \pm 1.09$ overlap between the hard samples identified by confidence- and energy-based methods. This overlap reduces to $45.67\% \pm 3.07$ when comparing straggler- and energy-based methods, and to $49.48\% \pm 3.04$ between straggler- and confidence-based methods. This explains the similarities between the results obtained via energy- and confidence-based methods, and the stark differences from the results obtained via straggler-based method.

\section{Benchmarking Procedure for Hard Sample Identification Methods}
\label{sec:benchmark}

\begin{figure}[t]
    \begin{subfigure}{7.5cm}
        \centering
        \includegraphics[width=7.5cm]{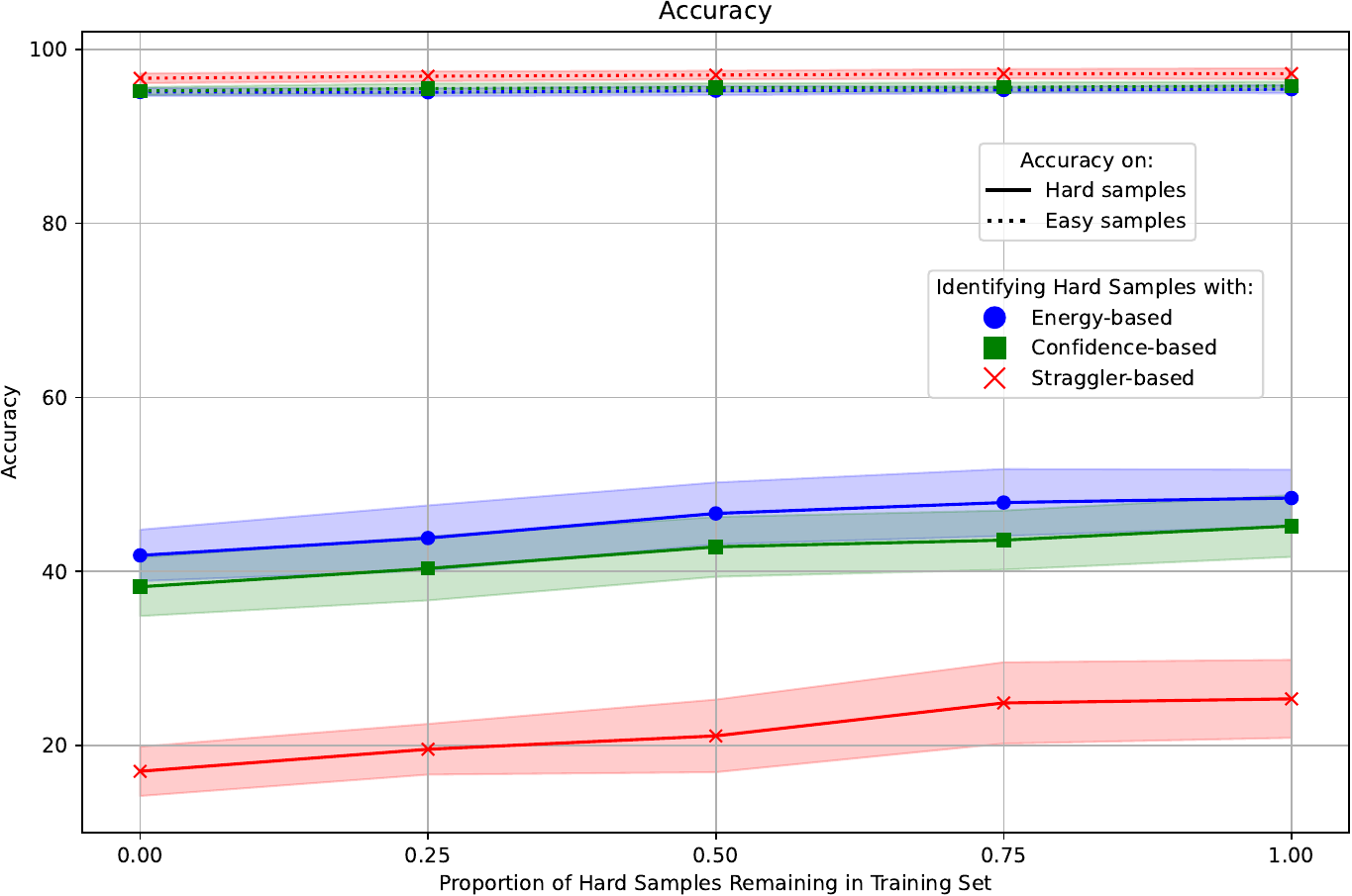}
        \caption{Increasing the number of hard samples in the training set}
    \end{subfigure}
    \begin{subfigure}{7.5cm}
        \centering
        \includegraphics[width=7.5cm]{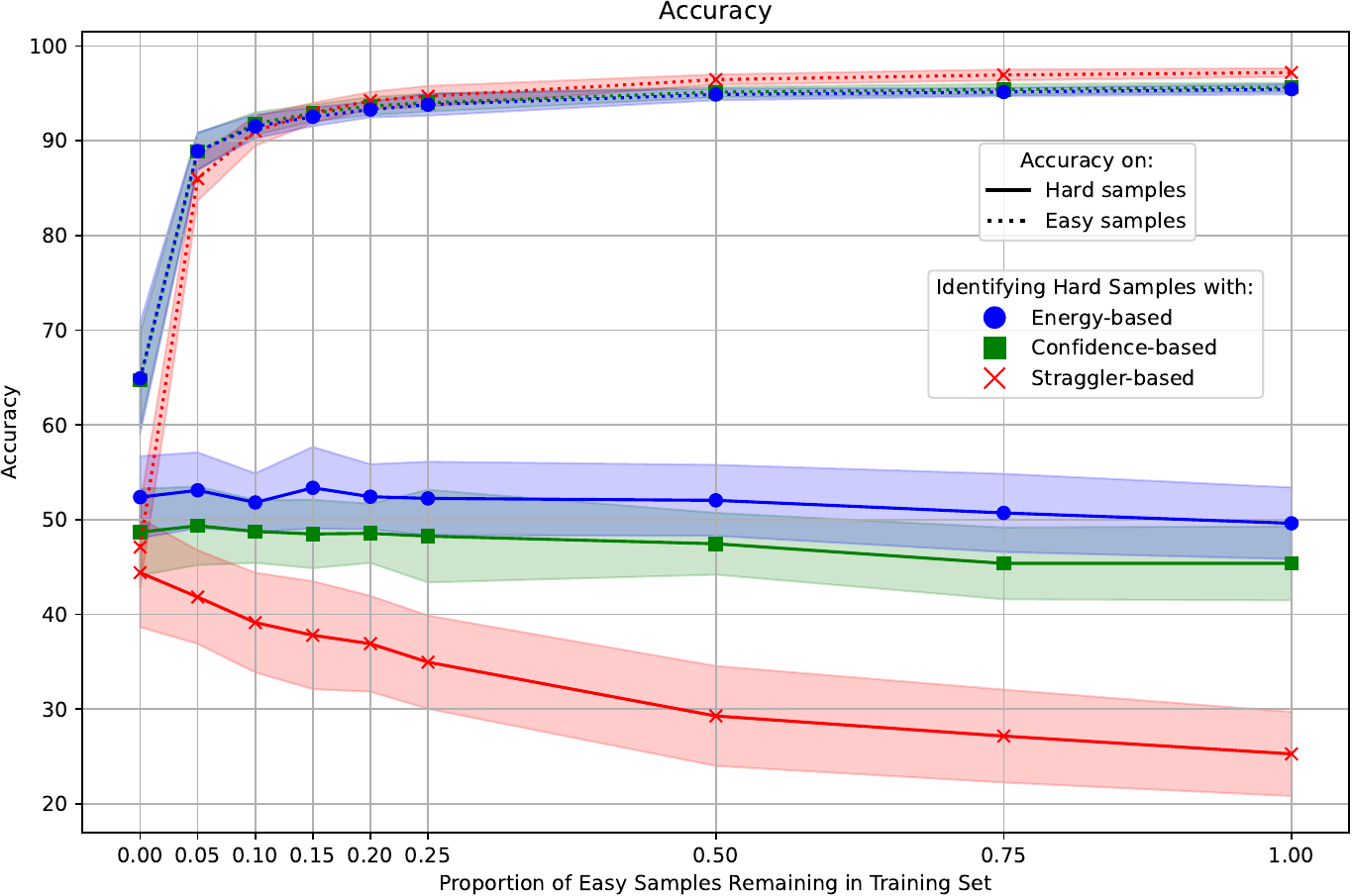}
        \caption{Increasing the number of easy samples in the training set}
    \end{subfigure}
    \caption{The in-class data imbalance becomes bigger when we combine the results of the performance on easy samples with those on hard samples in a single figure. It becomes clear that models achieve significantly better accuracy on easy samples than on hard samples, a discrepancy also observed in the between-class data imbalance between majority and minority classes. This in-class data imbalance can become either less or more pronounced depending on how successfully we manage to divide the dataset into hard and easy samples. The overlap between hard and easy sample sets, due to less precise identification, leads to a lower in-class data imbalance, as observed when comparing the accuracies of straggler-based methods with those of confidence-based or energy-based methods. At the top, green (confidence-based) and blue (energy-based) lines overlap each other at this scale.}
    \label{fig:fig5}
\end{figure}

\textbf{Motivation:} After identifying the existence of the in-class data imbalance our objective shifts to introducing a benchmarking procedure for the methods that identify hard samples. The knowledge from the between-class data imbalance suggests that the better the technique for identifying hard samples the larger the observed in-class data imbalance. This is because hard samples are harder to learn than easy samples, and we should see lower accuracy on hard samples than easy samples. To verify if this theory-based scenario happens in practise we explore both confidence-based and energy-based approaches as hard sample identifiers, as they are widely recognized for estimating sample difficulty in curriculum learning and for detecting anomalies. Our objective is to contrast these approaches with the more objective, dataset-level, straggler-based approach.

\noindent\textbf{Benchmarking setup:} Both confidence-based and energy-based methods employ a predetermined threshold to identify hard samples, clearly differentiating them from the straggler-based approach, which does not require such a threshold and thus avoids the bias introduced by it. However, to make the results more comparable, we need to establish this threshold using the straggler count. After determining the threshold, we train neural networks following the setup described in Sections \ref{sec:generalization_of_stragglers} and \ref{sec:strategic_sample_distribution}, and extract the most uncertain data samples using either confidence or energy as the metric of uncertainty. These 3 sets of hard and easy samples are then used to re-evaluate the degree of in-class data imbalance on MNIST, as illustrated in Fig. \ref{fig:fig5}.

\noindent\textbf{Strengths of straggler-based method:} After combining the results of the performance on easy samples with those on hard test samples in Figure \ref{fig:fig5}, the characteristics of in-class data imbalance become even more apparent. Across all identification methods used, we notice that: 1) models perform better on easy than on hard samples; and 2) to achieve the largest accuracy improvement for either easy or hard samples, we need to incorporate more of the corresponding samples into the training set. Both observations closely resemble the between-class data imbalance as thoroughly described in the previous section. However, we find that the degree of this in-class data imbalance can become less or more pronounced, depending on how we identify hard samples. For example, using confidence-based and energy-based approaches to identify hard samples makes the gap between the accuracy on easy and hard samples appear smaller than what the straggler-based method suggests. This implies that hard samples identified by the confidence-based and energy-based methods contain a high percentage of easy samples, making them appear to perform better on hard samples. Furthermore, using these methods makes it seem like training on hard samples alone allows the model to generalize to over 70\% onto easy samples. However, the clearest indication that confidence-based and energy-based methods are poor at identifying hard samples is the fact that they make it seem like adding easy samples to the training set, which constitute the majority of class manifolds, has a negligible effect on performance on hard samples, a conclusion completely contradictory to common knowledge from between-class data imbalance. Consequently, we posit that this variation in the degree of in-class data imbalance is crucial information that can be used to benchmark methods that identify hard samples. This idea stems from the resemblance of the in-class data imbalance to the between-class data imbalance thoroughly described in Sec. \ref{sec:strategic_sample_distribution}. The better we identify hard and easy samples, the more pronounced we can expect the in-class data imbalance to be.

\noindent\textbf{Distinction between hard samples and anomalies:} In our analysis, the expected performance disparity between confidence- and energy-based methods, well-documented in the context of out-of-distribution and anomaly detection \citep{liu2020energy}, does not manifest in the realm of hard sample identification. This observation underscores that residing in a low-density region does not inherently make a sample difficult to classify, nor does presence in a high-density area guarantee ease of classification. The essence of this insight lies in the divergence between anomaly detection, which often equates low density with being an anomaly, and the geometric complexities involved in learning from class manifolds. Samples located in geometrically simpler, yet sparsely populated areas of the manifold, may not present significant classification challenges, prompting us to reevaluate our understanding of what constitutes a hard sample in machine learning. This revelation underscores the necessity of distinguishing between anomalies and hard samples, advocating for methodologies that are attuned to the nuanced geometry of data distributions.


\section{Broader Impact Statement}

Our methodology emphasizes the importance of recognizing and addressing in-class data imbalance, fostering the development of AI systems that are more accurate and equitable. By enhancing model sensitivity to hard samples, we aim to reduce bias in AI predictions, underscoring that not all data are created equal. Focusing on hard samples for tasks such as fine-tuning and data augmentation could not only improve performance but also decrease computational complexity due to the reduced data size needed, thereby contributing to a reduction in the environmental impact associated with these tasks.

\section{Future Work}

Our work suggests directions for future research:
\begin{itemize}
    \item \textbf{Expanding to Imbalanced Datasets:} Generalizing our methods to imbalanced datasets is crucial, as accuracy is less informative there.
    \item \textbf{Clarifying Sample Hardness:} We posit that hard samples are in complex class manifold regions. A deeper investigation into the nature and origins of sample complexity is needed.
    \item \textbf{Leveraging Between-Class Tools:} Considering between-class imbalance as an unsupervised variant of in-class imbalance, adapting existing imbalance correction tools for in-class challenges warrants exploration.
\end{itemize}

\section{Conclusion and Limitations}

This study extends the foundational work of \citet{ciceri2024inversion}, illustrating that the dynamic radii changes of class manifolds, indicative of non-monotonous learning progression, persist in multiclass classification scenarios albeit with increased instability. Through the segmentation of MNIST, FashionMNIST, and KMNIST datasets into easy and hard samples, we reveal the pivotal role of hard sample distribution in modulating training and testing difficulty, thereby challenging the conventional reliance on test accuracy as a primary indicator of model generalization. Our investigation underscores the distinction between performance on easy versus hard samples. Finally, we introduce novel and intuitive benchmarking procedure for evaluating the effectiveness of various methods in identifying hard samples.

While our findings shed light on important aspects of sample difficulty and generalization, we acknowledge several limitations. The theoretical basis of our analysis heavily leans on the manifold hypothesis, which may not fully encompass the intricacies of data geometry in complex datasets. The simplicity of the models and datasets employed in our study, while facilitating a clearer understanding of foundational dynamics, may not accurately reflect the challenges present in more sophisticated machine learning tasks. Furthermore, the benchmarking procedure we introduce is not a comprehensive benchmark, but rather a set of guidelines that enable the comparison of hard sample identifiers.

In conclusion, this paper does not advocate for a specific solution but rather sheds light on the limitations of the widely utilized metric of test accuracy in ML and AutoML. It is our hope that subsequent research will be directed towards uncovering more effective methods for detecting hard samples and gaining a clearer insight into the disparate generalization types, ultimately leading to the development of models tailored to address these challenges.

\bibliography{paper}

\begin{thebibliography}{}

\bibitem[Ansuini et~al., 2019]{ansuini2019intrinsic}
Ansuini, A., Laio, A., Macke, J.~H., and Zoccolan, D. (2019).
\newblock Intrinsic dimension of data representations in deep neural networks.
\newblock {\em Advances in Neural Information Processing Systems}, 32.

\bibitem[Anwar et~al., 2014]{anwar2014measurement}
Anwar, N., Jones, G., and Ganesh, S. (2014).
\newblock Measurement of data complexity for classification problems with unbalanced data.
\newblock {\em Statistical Analysis and Data Mining: The ASA Data Science Journal}, 7(3):194--211.

\bibitem[Arpit et~al., 2017]{arpit2017closer}
Arpit, D., Jastrz{\k{e}}bski, S., Ballas, N., Krueger, D., Bengio, E., Kanwal, M.~S., Maharaj, T., Fischer, A., Courville, A., Bengio, Y., et~al. (2017).
\newblock A closer look at memorization in deep networks.
\newblock In {\em International conference on machine learning}, pages 233--242. PMLR.

\bibitem[Balestriero et~al., 2022]{balestriero2022effects}
Balestriero, R., Bottou, L., and LeCun, Y. (2022).
\newblock The effects of regularization and data augmentation are class dependent.
\newblock {\em Advances in Neural Information Processing Systems}, 35:37878--37891.

\bibitem[Bengio et~al., 2009]{bengio2009curriculum}
Bengio, Y., Louradour, J., Collobert, R., and Weston, J. (2009).
\newblock Curriculum learning.
\newblock In {\em Proceedings of the 26th annual international conference on machine learning}, pages 41--48.

\bibitem[Birdal et~al., 2021]{birdal2021intrinsic}
Birdal, T., Lou, A., Guibas, L.~J., and Simsekli, U. (2021).
\newblock Intrinsic dimension, persistent homology and generalization in neural networks.
\newblock {\em Advances in Neural Information Processing Systems}, 34:6776--6789.

\bibitem[Brendel et~al., 2017]{brendel2017decision}
Brendel, W., Rauber, J., and Bethge, M. (2017).
\newblock Decision-based adversarial attacks: Reliable attacks against black-box machine learning models.
\newblock {\em arXiv preprint arXiv:1712.04248}.

\bibitem[Budd et~al., 2021]{budd2021survey}
Budd, S., Robinson, E.~C., and Kainz, B. (2021).
\newblock A survey on active learning and human-in-the-loop deep learning for medical image analysis.
\newblock {\em Medical Image Analysis}, 71:102062.

\bibitem[Can{\'e}vet and Fleuret, 2015]{canevet2015efficient}
Can{\'e}vet, O. and Fleuret, F. (2015).
\newblock Efficient sample mining for object detection.
\newblock In {\em Asian Conference on Machine Learning}, pages 48--63. PMLR.

\bibitem[Chandola et~al., 2009]{chandola2009anomaly}
Chandola, V., Banerjee, A., and Kumar, V. (2009).
\newblock Anomaly detection: A survey.
\newblock {\em ACM computing surveys (CSUR)}, 41(3):1--58.

\bibitem[Chawla et~al., 2002]{chawla2002smote}
Chawla, N.~V., Bowyer, K.~W., Hall, L.~O., and Kegelmeyer, W.~P. (2002).
\newblock Smote: synthetic minority over-sampling technique.
\newblock {\em Journal of artificial intelligence research}, 16:321--357.

\bibitem[Cheung and Yeung, 2021]{cheung2021adaaug}
Cheung, T.-H. and Yeung, D.-Y. (2021).
\newblock Adaaug: Learning class-and instance-adaptive data augmentation policies.
\newblock In {\em International Conference on Learning Representations}.

\bibitem[Ciceri et~al., 2024]{ciceri2024inversion}
Ciceri, S., Cassani, L., Osella, M., Rotondo, P., Valle, F., and Gherardi, M. (2024).
\newblock Inversion dynamics of class manifolds in deep learning reveals tradeoffs underlying generalization.
\newblock {\em Nature Machine Intelligence}, pages 1--8.

\bibitem[Clanuwat et~al., 2018]{clanuwat2018deep}
Clanuwat, T., Bober-Irizar, M., Kitamoto, A., Lamb, A., Yamamoto, K., and Ha, D. (2018).
\newblock Deep learning for classical japanese literature.
\newblock {\em arXiv preprint arXiv:1812.01718}.

\bibitem[Cortes and Vapnik, 1995]{cortes1995support}
Cortes, C. and Vapnik, V. (1995).
\newblock Support-vector networks.
\newblock {\em Machine learning}, 20:273--297.

\bibitem[Cubuk et~al., 2018]{cubuk2018autoaugment}
Cubuk, E.~D., Zoph, B., Mane, D., Vasudevan, V., and Le, Q.~V. (2018).
\newblock Autoaugment: Learning augmentation policies from data.
\newblock {\em arXiv preprint arXiv:1805.09501}.

\bibitem[Deng, 2012]{deng2012mnist}
Deng, L. (2012).
\newblock The mnist database of handwritten digit images for machine learning research.
\newblock {\em IEEE Signal Processing Magazine}, 29(6):141--142.

\bibitem[Ericsson et~al., 2023]{ericsson2023better}
Ericsson, L., Li, D., and Hospedales, T. (2023).
\newblock Better practices for domain adaptation.
\newblock In {\em International Conference on Automated Machine Learning}, pages 4--1. PMLR.

\bibitem[Fefferman et~al., 2016]{fefferman2016testing}
Fefferman, C., Mitter, S., and Narayanan, H. (2016).
\newblock Testing the manifold hypothesis.
\newblock {\em Journal of the American Mathematical Society}, 29(4):983--1049.

\bibitem[Goldstein and Uchida, 2016]{goldstein2016comparative}
Goldstein, M. and Uchida, S. (2016).
\newblock A comparative evaluation of unsupervised anomaly detection algorithms for multivariate data.
\newblock {\em PloS one}, 11(4):e0152173.

\bibitem[Goodfellow et~al., 2014]{goodfellow2014explaining}
Goodfellow, I.~J., Shlens, J., and Szegedy, C. (2014).
\newblock Explaining and harnessing adversarial examples.
\newblock {\em arXiv preprint arXiv:1412.6572}.

\bibitem[Ho and Nvasconcelos, 2020]{ho2020contrastive}
Ho, C.-H. and Nvasconcelos, N. (2020).
\newblock Contrastive learning with adversarial examples.
\newblock {\em Advances in Neural Information Processing Systems}, 33:17081--17093.

\bibitem[Hodge and Austin, 2004]{hodge2004survey}
Hodge, V. and Austin, J. (2004).
\newblock A survey of outlier detection methodologies.
\newblock {\em Artificial intelligence review}, 22:85--126.

\bibitem[Holte et~al., 1989]{holte1989concept}
Holte, R.~C., Acker, L., Porter, B.~W., et~al. (1989).
\newblock Concept learning and the problem of small disjuncts.
\newblock In {\em IJCAI}, volume~89, pages 813--818.

\bibitem[Japkowicz, 2001]{japkowicz2001concept}
Japkowicz, N. (2001).
\newblock Concept-learning in the presence of between-class and within-class imbalances.
\newblock In {\em Advances in Artificial Intelligence: 14th Biennial Conference of the Canadian Society for Computational Studies of Intelligence, AI 2001 Ottawa, Canada, June 7--9, 2001 Proceedings 14}, pages 67--77. Springer.

\bibitem[Jo and Japkowicz, 2004]{jo2004class}
Jo, T. and Japkowicz, N. (2004).
\newblock Class imbalances versus small disjuncts.
\newblock {\em ACM Sigkdd Explorations Newsletter}, 6(1):40--49.

\bibitem[Kaufman and Azencot, 2023]{DBLP:conf/icml/KaufmanA23}
Kaufman, I. and Azencot, O. (2023).
\newblock Data representations' study of latent image manifolds.
\newblock In Krause, A., Brunskill, E., Cho, K., Engelhardt, B., Sabato, S., and Scarlett, J., editors, {\em International Conference on Machine Learning, {ICML} 2023, 23-29 July 2023, Honolulu, Hawaii, {USA}}, volume 202 of {\em Proceedings of Machine Learning Research}, pages 15928--15945. {PMLR}.

\bibitem[Kerssies, 2023]{kerssies2023neural}
Kerssies, T. (2023).
\newblock Neural architecture search for visual anomaly segmentation.
\newblock {\em arXiv preprint arXiv:2304.08975}.

\bibitem[Kienitz et~al., 2022]{kienitz2022effect}
Kienitz, D., Komendantskaya, E., and Lones, M. (2022).
\newblock The effect of manifold entanglement and intrinsic dimensionality on learning.
\newblock In {\em Proceedings of the AAAI Conference on Artificial Intelligence}, volume~36, pages 7160--7167.

\bibitem[Leevy et~al., 2018]{leevy2018survey}
Leevy, J.~L., Khoshgoftaar, T.~M., Bauder, R.~A., and Seliya, N. (2018).
\newblock A survey on addressing high-class imbalance in big data.
\newblock {\em Journal of Big Data}, 5(1):1--30.

\bibitem[Li et~al., 2021]{li2021autobalance}
Li, M., Zhang, X., Thrampoulidis, C., Chen, J., and Oymak, S. (2021).
\newblock Autobalance: Optimized loss functions for imbalanced data.
\newblock {\em Advances in Neural Information Processing Systems}, 34:3163--3177.

\bibitem[Li et~al., 2024]{li2024multi}
Li, S., Song, L., Wu, X., Hu, Z., Cheung, Y.-m., and Yao, X. (2024).
\newblock Multi-class imbalance classification based on data distribution and adaptive weights.
\newblock {\em IEEE Transactions on Knowledge and Data Engineering}.

\bibitem[Lim et~al., 2019]{lim2019fast}
Lim, S., Kim, I., Kim, T., Kim, C., and Kim, S. (2019).
\newblock Fast autoaugment.
\newblock {\em Advances in Neural Information Processing Systems}, 32.

\bibitem[Liu et~al., 2020]{liu2020energy}
Liu, W., Wang, X., Owens, J., and Li, Y. (2020).
\newblock Energy-based out-of-distribution detection.
\newblock {\em Advances in neural information processing systems}, 33:21464--21475.

\bibitem[Makarova et~al., 2022]{makarova2022automatic}
Makarova, A., Shen, H., Perrone, V., Klein, A., Faddoul, J.~B., Krause, A., Seeger, M., and Archambeau, C. (2022).
\newblock Automatic termination for hyperparameter optimization.
\newblock In {\em International Conference on Automated Machine Learning}, pages 7--1. PMLR.

\bibitem[Naitzat et~al., 2020]{naitzat2020topology}
Naitzat, G., Zhitnikov, A., and Lim, L.-H. (2020).
\newblock Topology of deep neural networks.
\newblock {\em The Journal of Machine Learning Research}, 21(1):7503--7542.

\bibitem[Narayanan and Niyogi, 2009]{narayanan2009sample}
Narayanan, H. and Niyogi, P. (2009).
\newblock On the sample complexity of learning smooth cuts on a manifold.
\newblock In {\em COLT}.

\bibitem[Northcutt et~al., 2021]{northcutt2021pervasive}
Northcutt, C.~G., Athalye, A., and Mueller, J. (2021).
\newblock Pervasive label errors in test sets destabilize machine learning benchmarks.
\newblock {\em arXiv preprint arXiv:2103.14749}.

\bibitem[Paszke et~al., 2019]{paszke2019pytorch}
Paszke, A., Gross, S., Massa, F., Lerer, A., Bradbury, J., Chanan, G., Killeen, T., Lin, Z., Gimelshein, N., Antiga, L., et~al. (2019).
\newblock Pytorch: An imperative style, high-performance deep learning library.
\newblock {\em Advances in neural information processing systems}, 32.

\bibitem[Pope et~al., 2021]{pope2021intrinsic}
Pope, P., Zhu, C., Abdelkader, A., Goldblum, M., and Goldstein, T. (2021).
\newblock The intrinsic dimension of images and its impact on learning.
\newblock {\em arXiv preprint arXiv:2104.08894}.

\bibitem[Ren et~al., 2021]{ren2021survey}
Ren, P., Xiao, Y., Chang, X., Huang, P.-Y., Li, Z., Gupta, B.~B., Chen, X., and Wang, X. (2021).
\newblock A survey of deep active learning.
\newblock {\em ACM computing surveys (CSUR)}, 54(9):1--40.

\bibitem[Serban et~al., 2020]{serban2020adversarial}
Serban, A., Poll, E., and Visser, J. (2020).
\newblock Adversarial examples on object recognition: A comprehensive survey.
\newblock {\em ACM Computing Surveys (CSUR)}, 53(3):1--38.

\bibitem[Settles, 2009]{settles2009active}
Settles, B. (2009).
\newblock Active learning literature survey.

\bibitem[Shrivastava et~al., 2016]{shrivastava2016training}
Shrivastava, A., Gupta, A., and Girshick, R. (2016).
\newblock Training region-based object detectors with online hard example mining.
\newblock In {\em Proceedings of the IEEE conference on computer vision and pattern recognition}, pages 761--769.

\bibitem[Soviany et~al., 2022]{soviany2022curriculum}
Soviany, P., Ionescu, R.~T., Rota, P., and Sebe, N. (2022).
\newblock Curriculum learning: A survey.
\newblock {\em International Journal of Computer Vision}, 130(6):1526--1565.

\bibitem[Wang et~al., 2021]{wang2021survey}
Wang, X., Chen, Y., and Zhu, W. (2021).
\newblock A survey on curriculum learning.
\newblock {\em IEEE Transactions on Pattern Analysis and Machine Intelligence}, 44(9):4555--4576.

\bibitem[Xiao et~al., 2017]{xiao2017fashion}
Xiao, H., Rasul, K., and Vollgraf, R. (2017).
\newblock Fashion-mnist: a novel image dataset for benchmarking machine learning algorithms.
\newblock {\em arXiv preprint arXiv:1708.07747}.

\bibitem[Yang et~al., 2023]{yang2023dcdetector}
Yang, Y., Zhang, C., Zhou, T., Wen, Q., and Sun, L. (2023).
\newblock Dcdetector: Dual attention contrastive representation learning for time series anomaly detection.
\newblock {\em arXiv preprint arXiv:2306.10347}.

\bibitem[Zhou and Wu, 2023]{zhou2023samples}
Zhou, X. and Wu, O. (2023).
\newblock Which samples should be learned first: Easy or hard?
\newblock {\em IEEE Transactions on Neural Networks and Learning Systems}.

\end{thebibliography}

\begin{acknowledgements}

This work was supported by the UK Research and Innovation (UKRI) Engineering and Physical Sciences Research Council (EPSRC) under the PhD Scholarship Grant.

\end{acknowledgements}






\newpage 
\section*{Submission Checklist}

\begin{enumerate}
\item For all authors\dots
  \begin{enumerate}
  \item Do the main claims made in the abstract and introduction accurately
    reflect the paper's contributions and scope?
    Yes
  \item Did you describe the limitations of your work?
    Yes
  \item Did you discuss any potential negative societal impacts of your work?
    No, but we don't think it's necessary due to the theoretical nature of our work
  \item Did you read the ethics review guidelines and ensure that your paper
    conforms to them? \url{https://2022.automl.cc/ethics-accessibility/}
    Yes
  \end{enumerate}
\item If you ran experiments\dots
  \begin{enumerate}
  \item Did you use the same evaluation protocol for all methods being compared (e.g., 
    same benchmarks, data (sub)sets, available resources)? 
    Yes. All experiments were run using the same architecture, and the same hyperparameters. We describe all experimental settings in Appendix B and D
  \item Did you specify all the necessary details of your evaluation (e.g., data splits,
    pre-processing, search spaces, hyperparameter tuning)?
    Yes
  \item Did you repeat your experiments (e.g., across multiple random seeds or splits) to account for the impact of randomness in your methods or data?
    Yes
  \item Did you report the uncertainty of your results (e.g., the variance across random seeds or splits)?
    Yes, the standard deviation is visible in the figures
  \item Did you report the statistical significance of your results?
    No. We have not conducted any tests to measure the statistical significance of our results.
  \item Did you use tabular or surrogate benchmarks for in-depth evaluations?
    Not applicable here
  \item Did you compare performance over time and describe how you selected the maximum duration?
    Not applicable here
  \item Did you include the total amount of compute and the type of resources
    used (e.g., type of \textsc{gpu}s, internal cluster, or cloud provider)?
    Yes, we have included the used resources and the approximate amount of time in Appendix \ref{sec:Replicating_inversion_point_appendix} and \ref{sec:verifying_importance_of_stragglers_appendix}.
  \item Did you run ablation studies to assess the impact of different
    components of your approach?
    No. We have not tried to see the impact of different architectures on our work. Similarly, we have not run an ablation study on the index of the latent layer used to compute the radii of class manifolds.
  \end{enumerate}
\item With respect to the code used to obtain your results\dots
  \begin{enumerate}
\item Did you include the code, data, and instructions needed to reproduce the
    main experimental results, including all requirements (e.g.,
    \texttt{requirements.txt} with explicit versions), random seeds, an instructive
    \texttt{README} with installation, and execution commands (either in the
    supplemental material or as a \textsc{url})?
    Yes
  \item Did you include a minimal example to replicate results on a small subset
    of the experiments or on toy data?
    Yes
  \item Did you ensure sufficient code quality and documentation so that someone else 
    can execute and understand your code?
    We hope we did.
  \item Did you include the raw results of running your experiments with the given
    code, data, and instructions?
    No
  \item Did you include the code, additional data, and instructions needed to generate
    the figures and tables in your paper based on the raw results?
    The code provided is enough to generate the figures from this paper.
  \end{enumerate}
\item If you used existing assets (e.g., code, data, models)\dots
  \begin{enumerate}
  \item Did you cite the creators of used assets?
    Yes
  \item Did you discuss whether and how consent was obtained from people whose
    data you're using/curating if the license requires it?
    No
  \item Did you discuss whether the data you are using/curating contains
    personally identifiable information or offensive content?
    Not applicable
  \end{enumerate}
\item If you created/released new assets (e.g., code, data, models)\dots
  \begin{enumerate}
    \item Did you mention the license of the new assets (e.g., as part of your code submission)?
    Yes
    \item Did you include the new assets either in the supplemental material or as
    a \textsc{url} (to, e.g., GitHub or Hugging Face)?
    Yes. We provided link to anonymous github repository.
  \end{enumerate}
\item If you used crowdsourcing or conducted research with human subjects\dots
  \begin{enumerate}
  \item Did you include the full text of instructions given to participants and
    screenshots, if applicable?
    N/A
  \item Did you describe any potential participant risks, with links to
    Institutional Review Board (\textsc{irb}) approvals, if applicable?
    N/A
  \item Did you include the estimated hourly wage paid to participants and the
    total amount spent on participant compensation?
    N/A
  \end{enumerate}
\item If you included theoretical results\dots
  \begin{enumerate}
  \item Did you state the full set of assumptions of all theoretical results?
    Not applicable
  \item Did you include complete proofs of all theoretical results?
    No proofs are necessary in our paper
  \end{enumerate}
\end{enumerate}

\newpage
\appendix

\section{Radii of Class Manifolds}
\label{sec:radii_description_appendix}

Throughout our paper, we have utilized the squared gyration radii of the class manifolds to identify the inversion points. We have adapted the formula from \citet{ciceri2024inversion} for the multiclass setting as follows:
\[R^2_i(t) = \frac{1}{2n^2_i} \sum_{x, y\in \mathcal{M}_i(t)} \|x - y\| ^2, \]
where $\mathcal{M}_i(t)$ is the internal representation of the manifold of class $i$ at epoch $t$, and $n_i = |\mathcal{M}_i(t)|$ is the number of elements from class $i$ (for all $k$ labels in the dataset). This metric effectively captures the essence of intra-class variability by summing the distances between all point pairs in the feature space, with a larger squared gyration radius indicating more dispersed points.

\section{Delving into Experiment from Section \ref{sec:generalization_of_stragglers}}
\label{sec:Replicating_inversion_point_appendix}

For training, we use FC ReLU NN with a configuration $784-20-20-10$, where $784$ is the input dimensionality for MNIST, KMNIST, and FashionMNIST datasets. The radii of class manifolds are computed using their latent representations from the first hidden layer. For training, we employ the Adam optimizer with a learning rate of $0.001$ and weight decay of $0.01$, alongside cross-entropy loss. The training is conducted over $500$ epochs. When performing experiments on the entire dataset, we combine the train and test sets, shuffle the obtained full dataset, and normalize it before training. For subset analyses, we extract a subset after shuffling and then normalize this subset. We have added a \texttt{compute\_radii} flag to expedite training when radii computation is not required (see line \ref{line8} of the pseudocode below for reference). Radii computation starts after epoch $20$, as results prior to that are highly chaotic, a phenomenon also observed by \citet{ciceri2024inversion}. Figures \ref{fig:fig2} and \ref{fig:fig7} represent the mean and standard deviation over $100$ runs. Our implementation is based on PyTorch \citep{paszke2019pytorch}. The experiments were run on a 12-core CPU of a MacBook Pro, taking $3-4$ hours on each dataset.

\begin{algorithm}
    \caption{Replicate Inversion Results on Multiple Classes}
    \begin{algorithmic}[1]
        \State \textbf{Input:} model, loader, optimizer, compute\_radii
        \State \textbf{Output:} epoch\_radii
        \State Initialize epoch\_radii as an empty list
        \For{each epoch from 1 to 500}
            \For{each batch (data, target) in loader}
                \State Perform forward and backpropagation
            \EndFor
            \If{compute\_radii is true AND epoch $>$ 20} \label{line8}
                \State Compute current\_radii of each for the class manifolds
                \State Append tuple (epoch, current\_radii) to epoch\_radii
            \EndIf
        \EndFor
        \State \textbf{return} epoch\_radii
    \end{algorithmic}
\end{algorithm}

\section{Generalization of Stragglers to Multiclass Classification on Other Datasets}
\label{sec:generalization_of_stragglers_appendix}

\begin{figure}[t!]
    \begin{subfigure}{12cm}
        \centering
        \includegraphics[width=12cm]{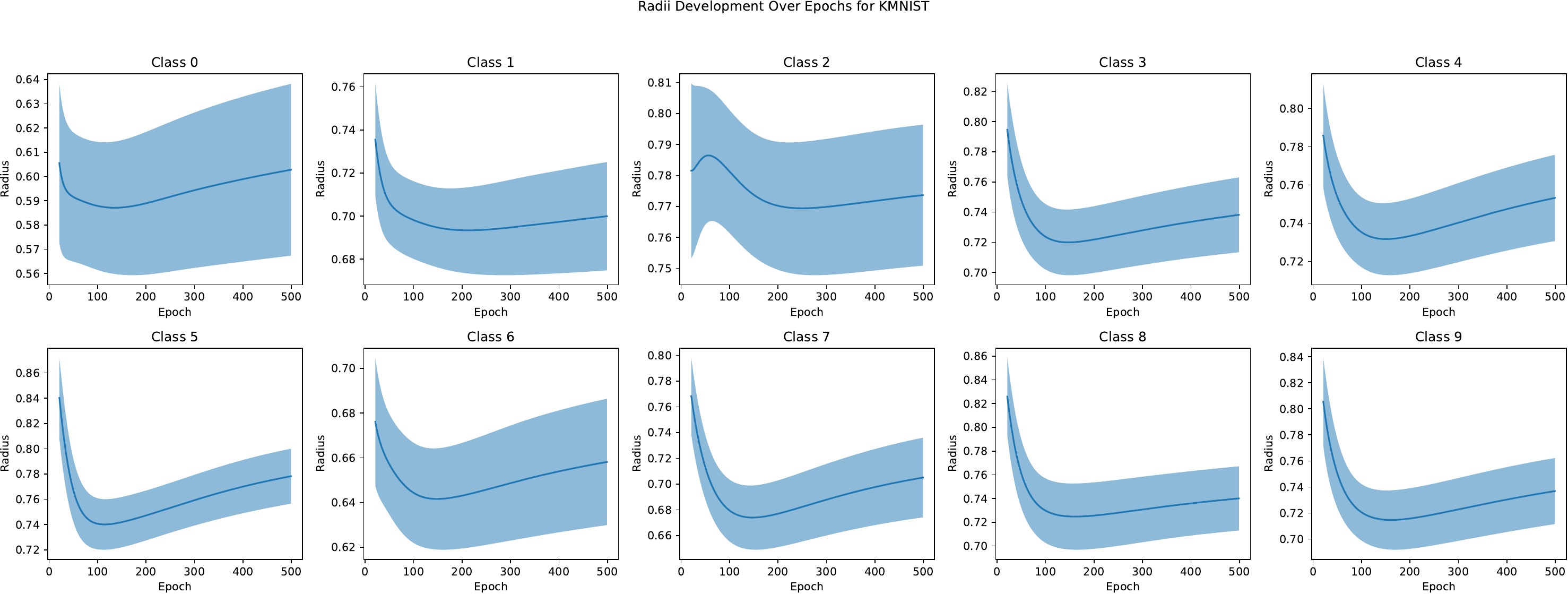}
        \caption{}
        \label{fig:radii_over_error_KMNIST}
    \end{subfigure}
    \centering
    \begin{subfigure}{12cm}
        \centering
        \includegraphics[width=12cm]{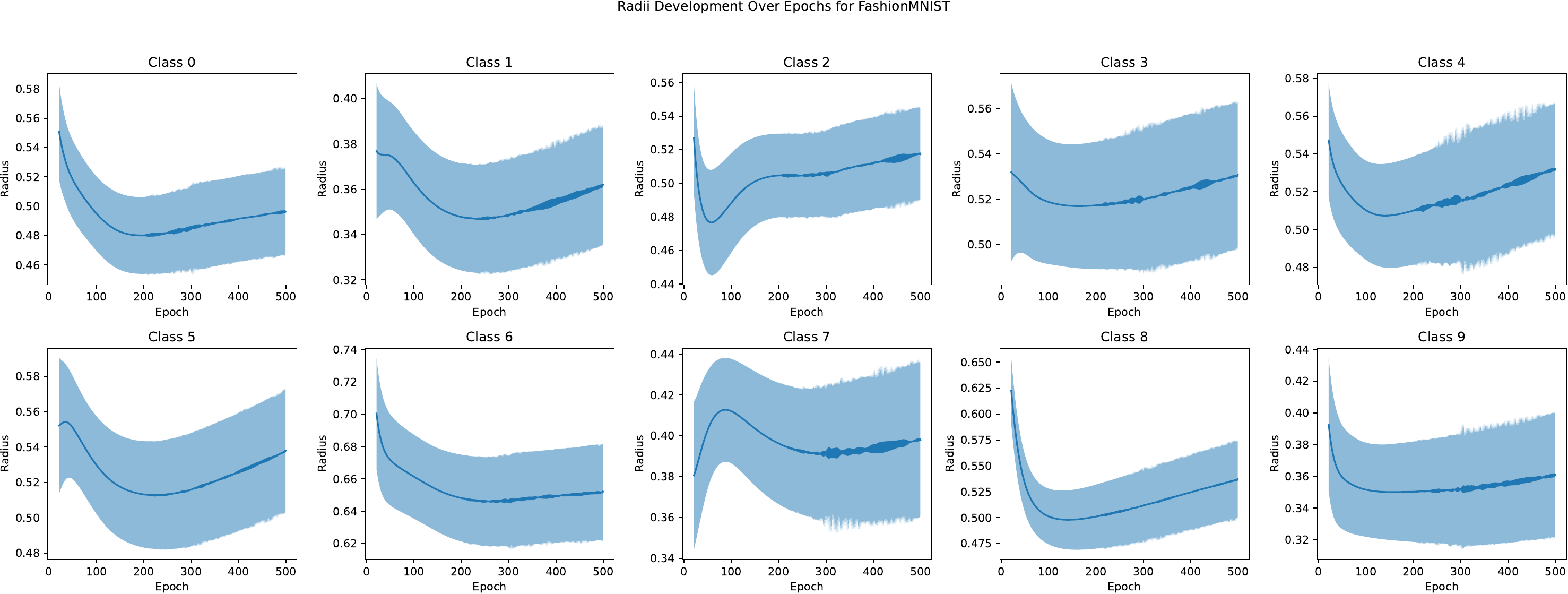}
        \caption{}
        \label{fig:radii_over_error_FashionMNIST}
    \end{subfigure}
    \caption{We see the same phenomena on KMNIST (a), and FashionMNIST (b), as we did in Sec. \ref{sec:generalization_of_stragglers}, with the radii behaving very chaotically during the later stages of learning on FashionMNIST (most likely due to it's increased complexity in comparison to MNIST and KMNIST).}
    \label{fig:fig7}
\end{figure}

We repeated the experiments from Section \ref{sec:generalization_of_stragglers} on FashionMNIST and KMNIST to validate whether the inversion point observed in our previous studies also appears when working with other datasets (see Figure \ref{fig:fig7}). The results confirm that the phenomena observed in Section \ref{sec:generalization_of_stragglers} are not exclusive to the MNIST dataset. An interesting difference, however, is observed upon inspecting the results on FashionMNIST: we noticed that the radii start to behave chaotically after a certain period of training. We hypothesize that this behavior stems from the increased complexity of the FashionMNIST dataset; however, further experiments are required to conclusively determine the cause of this trend.

In their work, \citet{ciceri2024inversion} presented evidence of the inversion point phenomenon using both epoch and test error as variables. However, unlike their approach, we conducted our experiments across the entirety of the datasets. Our aim was not to discover the inversion point phenomenon but rather to utilize it for identifying all hard samples within the datasets. Consequently, we report our results using only epoch as a variable.

\section{Delving into Experiment from Section \ref{sec:strategic_sample_distribution}}
\label{sec:verifying_importance_of_stragglers_appendix}

We employ the same experimental setup as described in Appendix \ref{sec:Replicating_inversion_point_appendix}. Our aim is to ensure that our results are invariant to model initialization and the obtained straggler set. Therefore, we generate three straggler sets, and for each of these sets, we measure the generalization with three networks, each with a distinct initialization. We then compute the mean and standard deviation over these nine runs to obtain Figures \ref{fig:fig3} and \ref{fig:fig4}. To identify a straggler set, we train a model and compute the radii of class manifolds after every epoch. Upon detecting an inversion point for a specific class, we save the parameters into a dictionary. This process results in $k$ models, one for each class (see pseudocode below), that define $k$ class-level straggler sets. These are later aggregated to form dataset-level stragglers. If an inversion point is not found within $500$ epochs, we rerun the \texttt{train\_stop\_at\_inversion} with a new model initialization. The experiments were conducted on NVIDIA A100 GPUs and took between $4-8$ hours when working on subsets consisting of $20,000$ samples and performing $25$ runs - $5$ straggler sets; $5$ initializations per straggler set (Figures \ref{fig:fig5}, \ref{fig:8} and \ref{fig:final_figure}). The computation time increases to approximately $8-16$ hours when running the experiments on all $70,000$ samples, but with only $9$ runs - $3$ straggler sets; $3$ initializations per straggler set (Figures \ref{fig:fig3} and \ref{fig:fig4}).

\begin{algorithm}
    \caption{Train Stop at Inversion}
    \begin{algorithmic}[1]
        \State \textbf{Input:} model, loader, optimizer
        \State \textbf{Output:} models, inversion\_points
        
        \State Initialize prev\_radii with infinity for each class index
        \State Initialize models as an empty dictionary
        \State Initialize found\_classes as an empty set
        \State Initialize inversion\_points as an empty dictionary
        
        \For{each epoch from 1 to 500}
            \For{each batch (data, target) in loader}
                \State Perform forward and backpropagation
            \EndFor
            \If{epoch modulo 5 is 0}
                \State Compute current\_radii of each for the class manifolds
                \For{each class index key in current\_radii}
                    \If{key not in models AND current\_radii[key] $>$ prev\_radii[key] AND epoch $>$ 20}
                        \State Save current model state to models at key
                        \State Add key to found\_classes
                        \State Record epoch as inversion point for key
                    \EndIf
                \EndFor
                \State Update prev\_radii with current\_radii
            \EndIf
            \If{all class manifolds have found models}
                \State Break the loop
            \EndIf
        \EndFor
        \State \textbf{return} models, inversion\_points
    \end{algorithmic}
\end{algorithm}

\section{Investigating Other Metrics and Hard Samples}
\label{sec:f1_investigation_appendix}

In this Section we attach the results of further experiments that didn't make it to the main text due to the page limit.

\begin{figure}[t!]
    \begin{subfigure}{5cm}
        \centering
        \includegraphics[width=5cm]{Figures/easy_accuracy.pdf}
        \caption{Accuracy vs the ratio of easy training samples}
    \end{subfigure}
    \centering
    \begin{subfigure}{5cm}
        \centering
        \includegraphics[width=5cm]{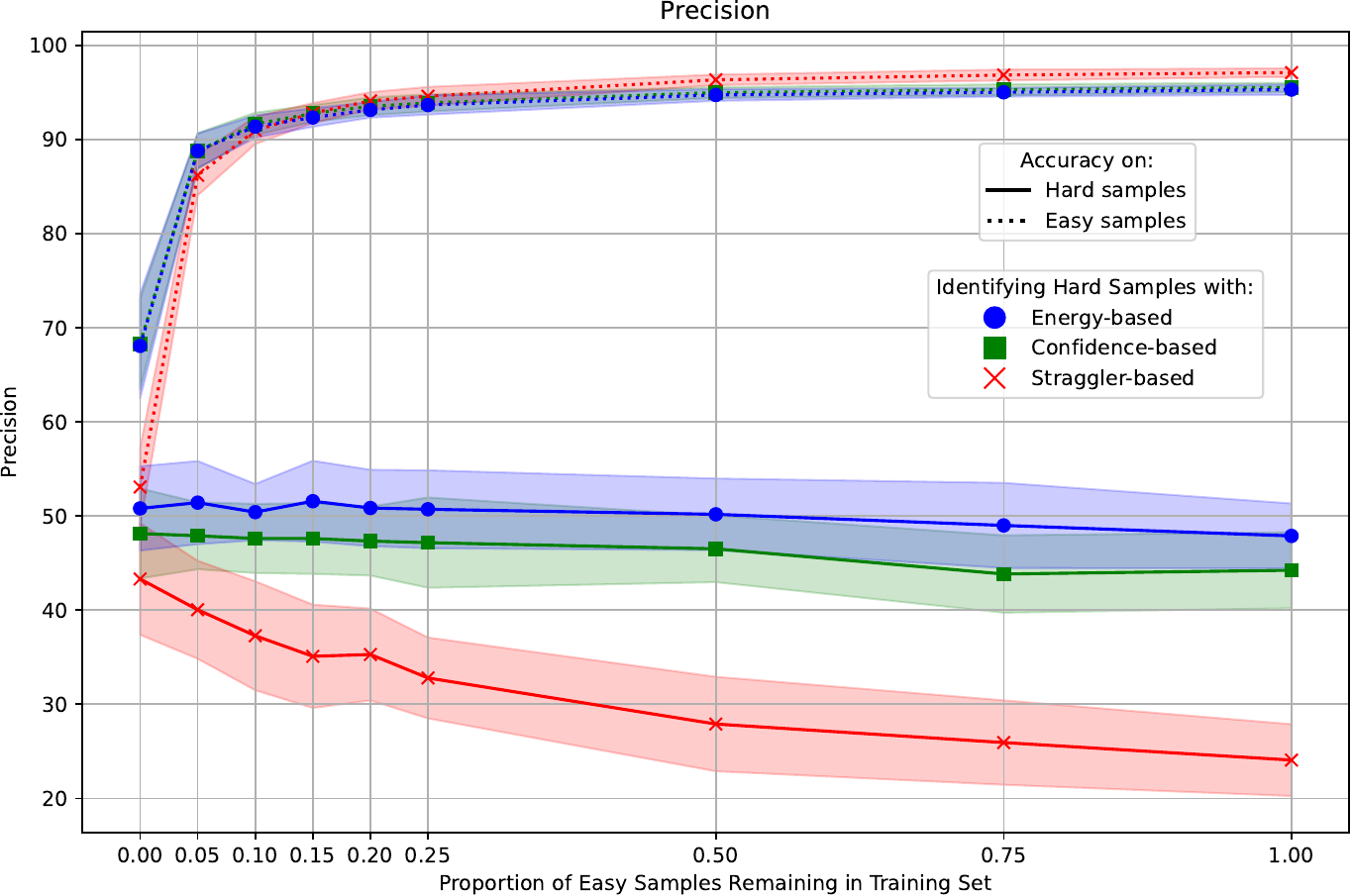}
        \caption{Recall vs the ratio of easy training samples}
    \end{subfigure}
    \centering
    \begin{subfigure}{5cm}
        \centering
        \includegraphics[width=5cm]{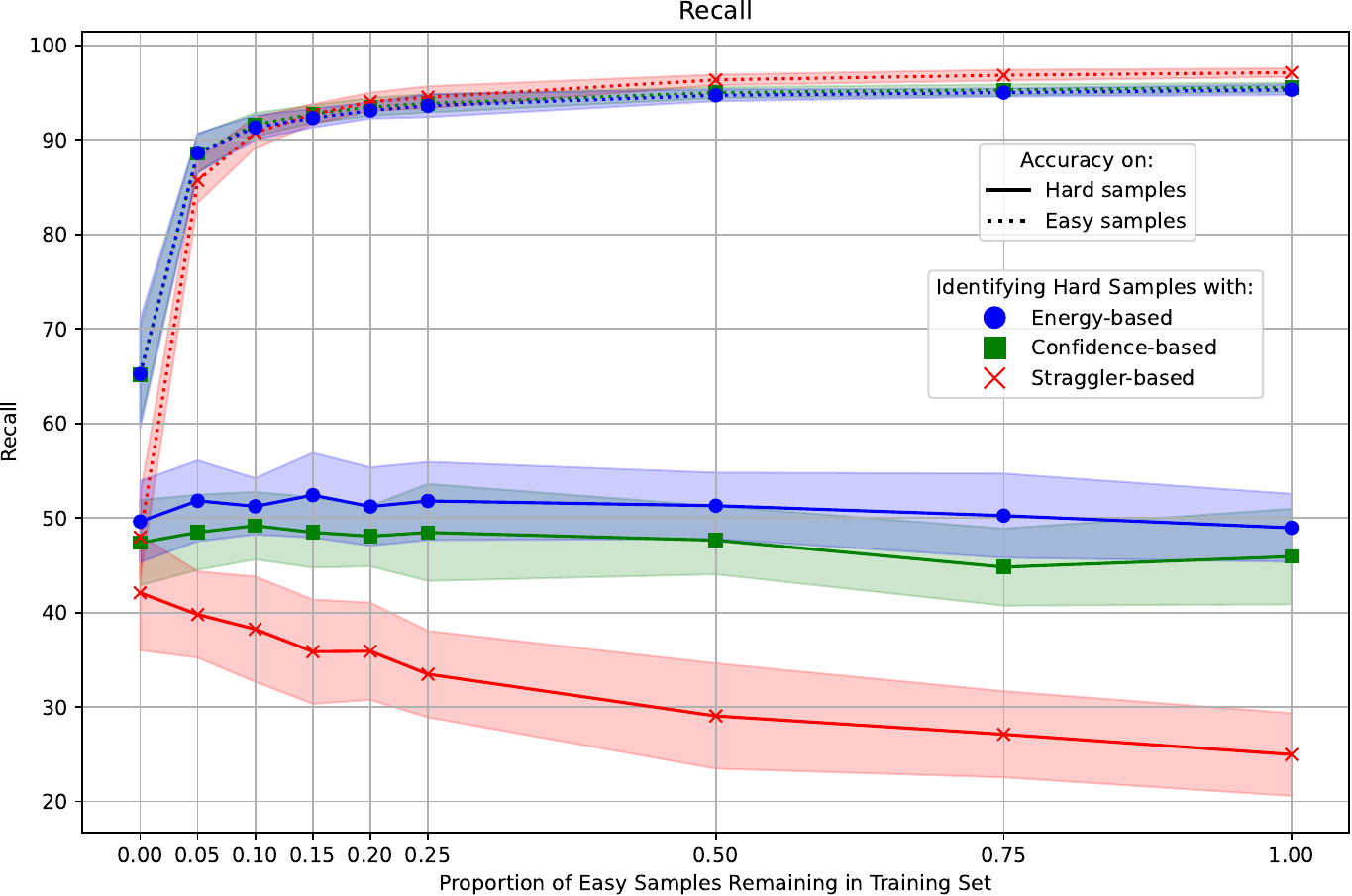}
        \caption{Precision vs the ratio of easy training samples}
    \end{subfigure}
    \centering
    \begin{subfigure}{5cm}
        \centering
        \includegraphics[width=5cm]{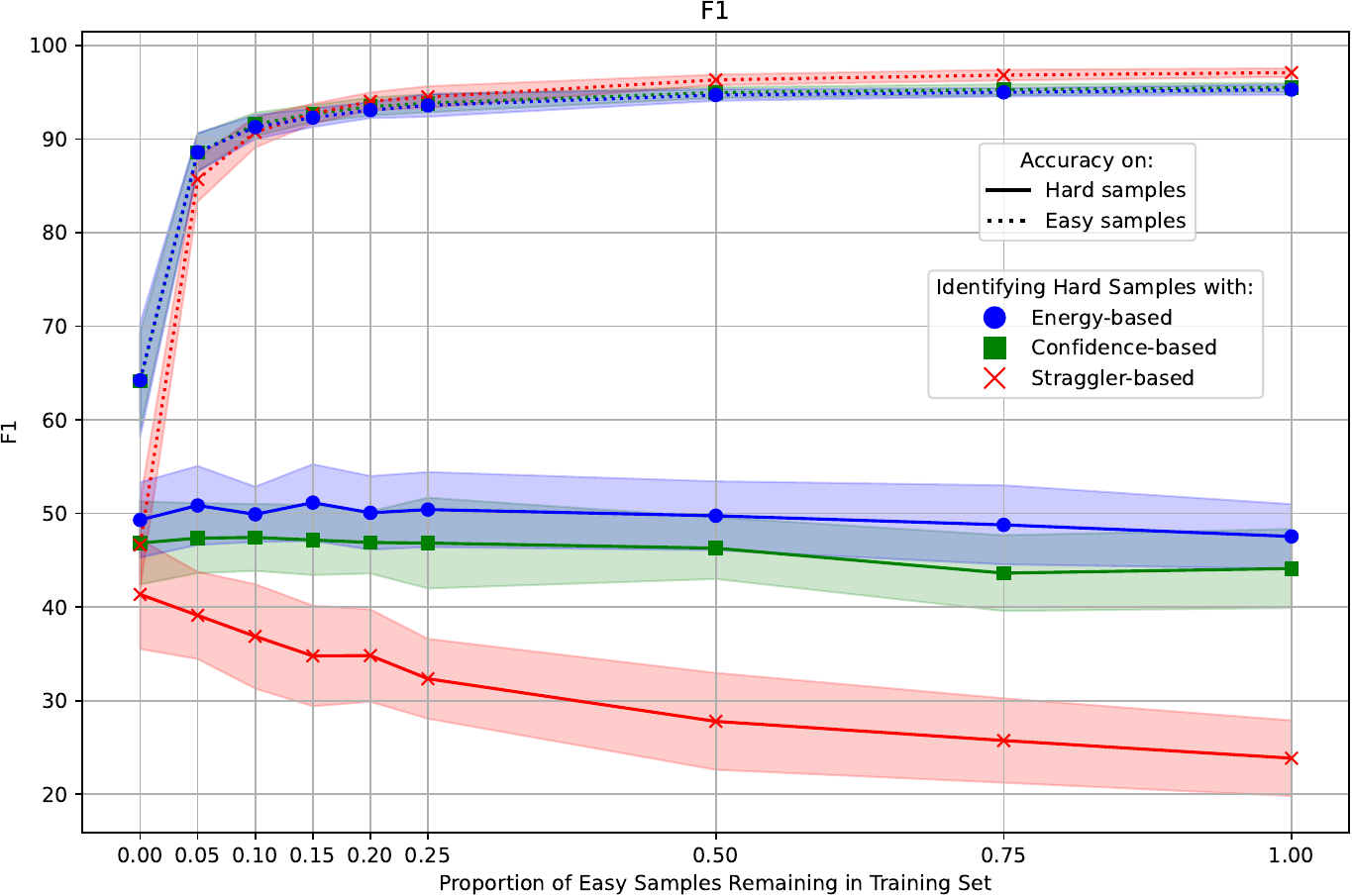}
        \caption{F1-Score vs the ratio of easy training samples}
    \end{subfigure}
    \begin{subfigure}{5cm}
        \centering
        \includegraphics[width=5cm]{Figures/hard_accuracy.pdf}
        \caption{Accuracy vs the ratio of hard training samples}
    \end{subfigure}
    \centering
    \begin{subfigure}{5cm}
        \centering
        \includegraphics[width=5cm]{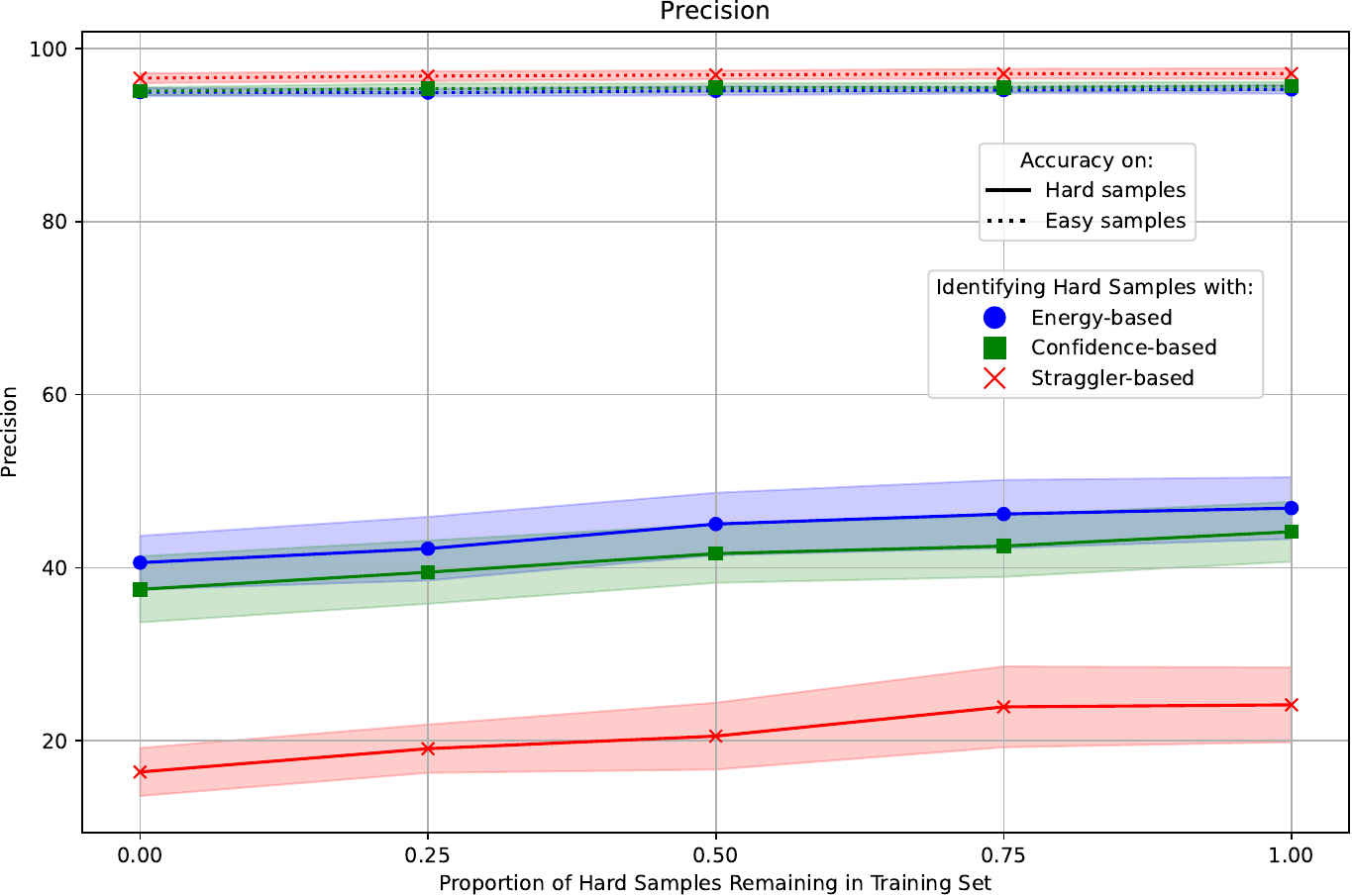}
        \caption{Recall vs the ratio of hard training samples}
    \end{subfigure}
    \centering
    \begin{subfigure}{5cm}
        \centering
        \includegraphics[width=5cm]{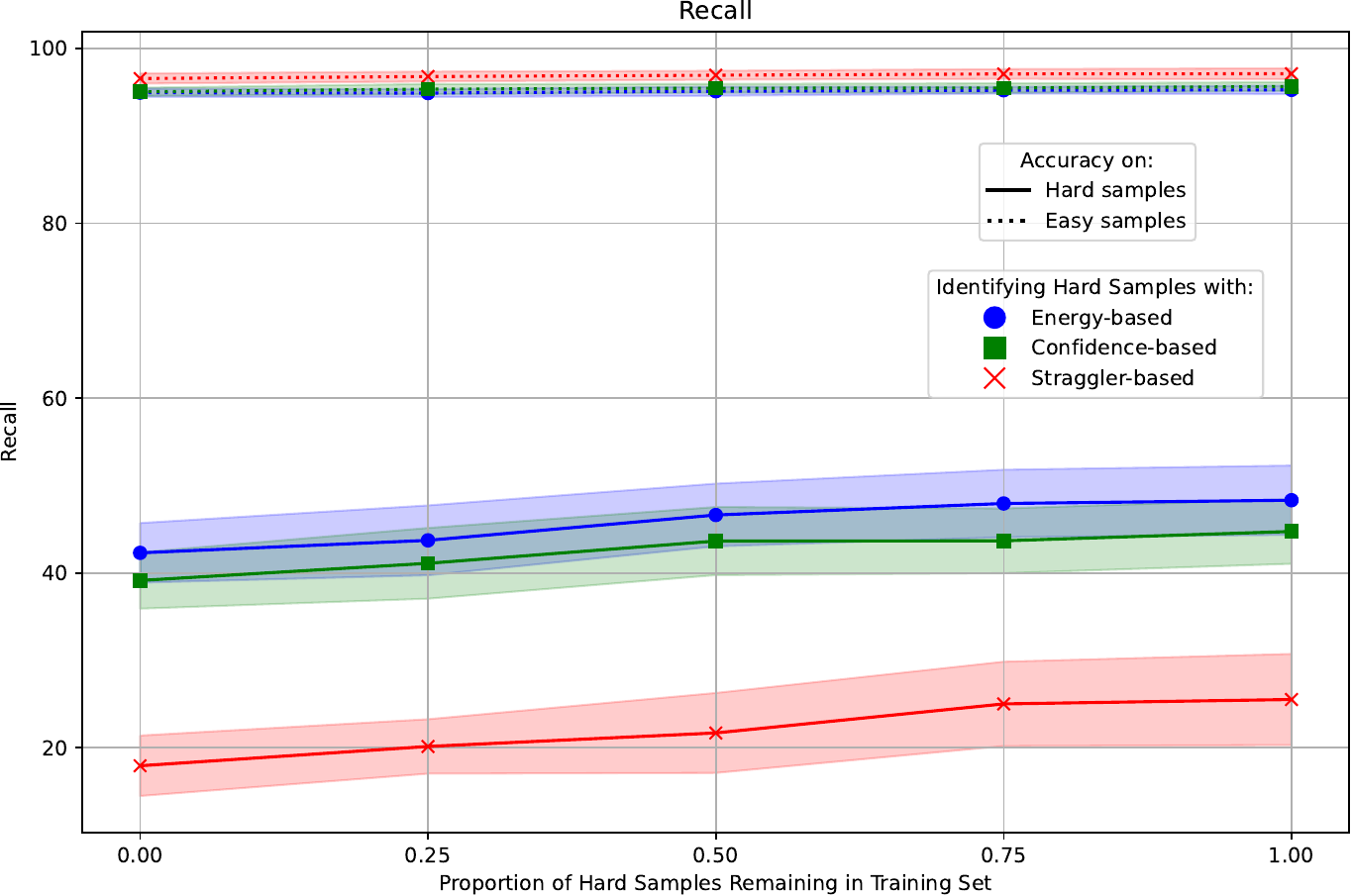}
        \caption{Precision vs the ratio of hard training samples}
    \end{subfigure}
    \centering
    \begin{subfigure}{5cm}
        \centering
        \includegraphics[width=5cm]{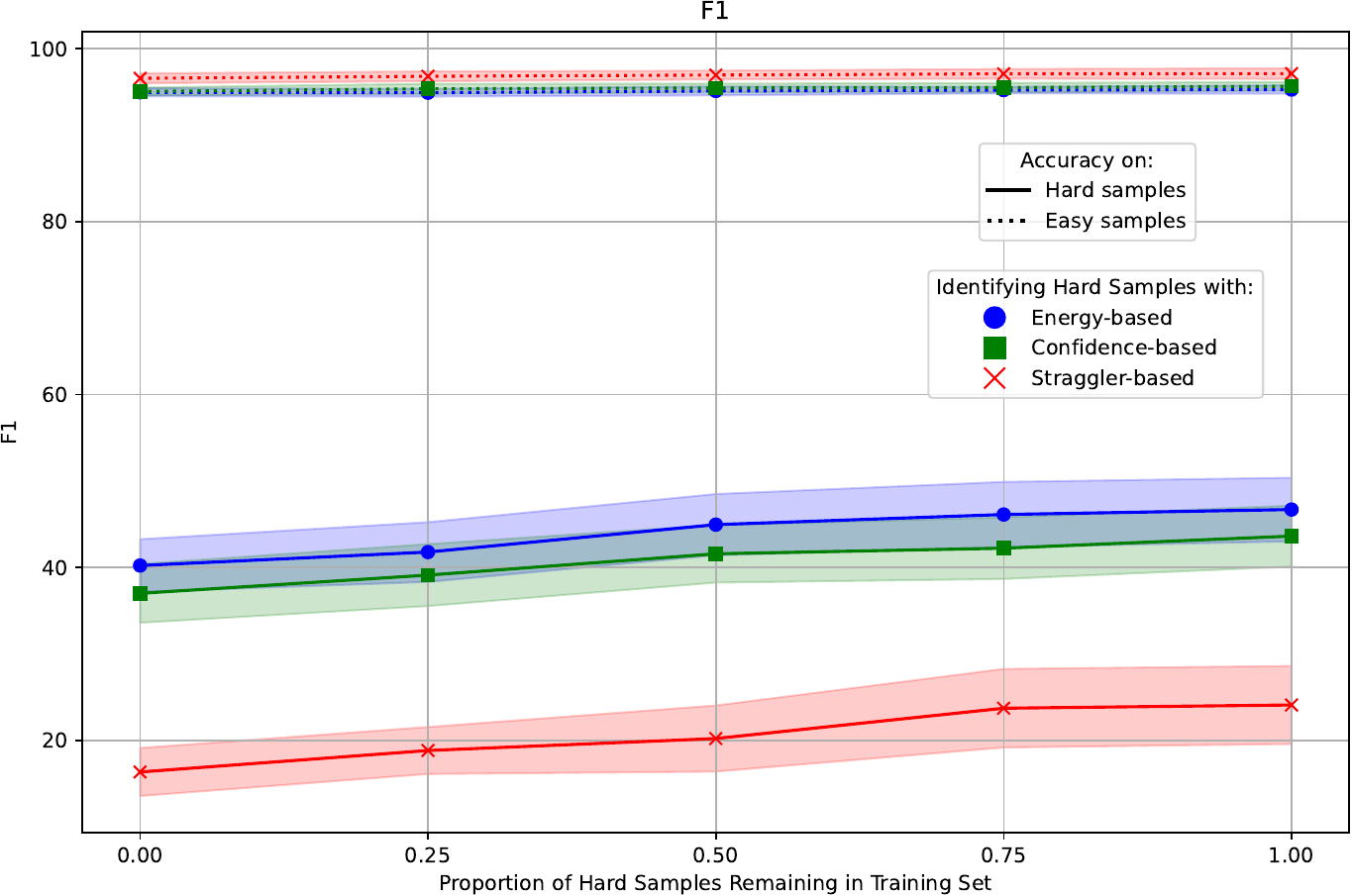}
        \caption{F1-Score vs the ratio of hard training samples}
    \end{subfigure}
    \caption{We reran the experiments from Section \ref{sec:strategic_sample_distribution}, this time also computing precision, recall, and F1 Score. We discovered that the metrics typically used to address between-class data imbalance are ineffective for investigating in-class data imbalance. We observed that both asymptotic precision and recall exhibit similar values. Figures a-d (e-h) display the results of modifying the ratio of easy (hard) samples in the training set.}
    \label{fig:8}
\end{figure}

As described in Section \ref{sec:strategic_sample_distribution}, we first measure the value of the additional information obtained by switching from accuracy to precision, recall and F1-score. Due to the balanced nature of the used datasets we do not see any additional valuable information, as all the metrics are closely correlated to each other. We can see that when looking at Figure \ref{fig:8}. However, this would definitely not be the case if we moved to class-imbalanced datasets. Over there we expect that accuracy would no longer grant us enough information to be able to apply our benchmarking procedure without incorporating F1-score.

\begin{figure}[t!]
    \begin{subfigure}{5cm}
        \centering
        \includegraphics[width=5cm]{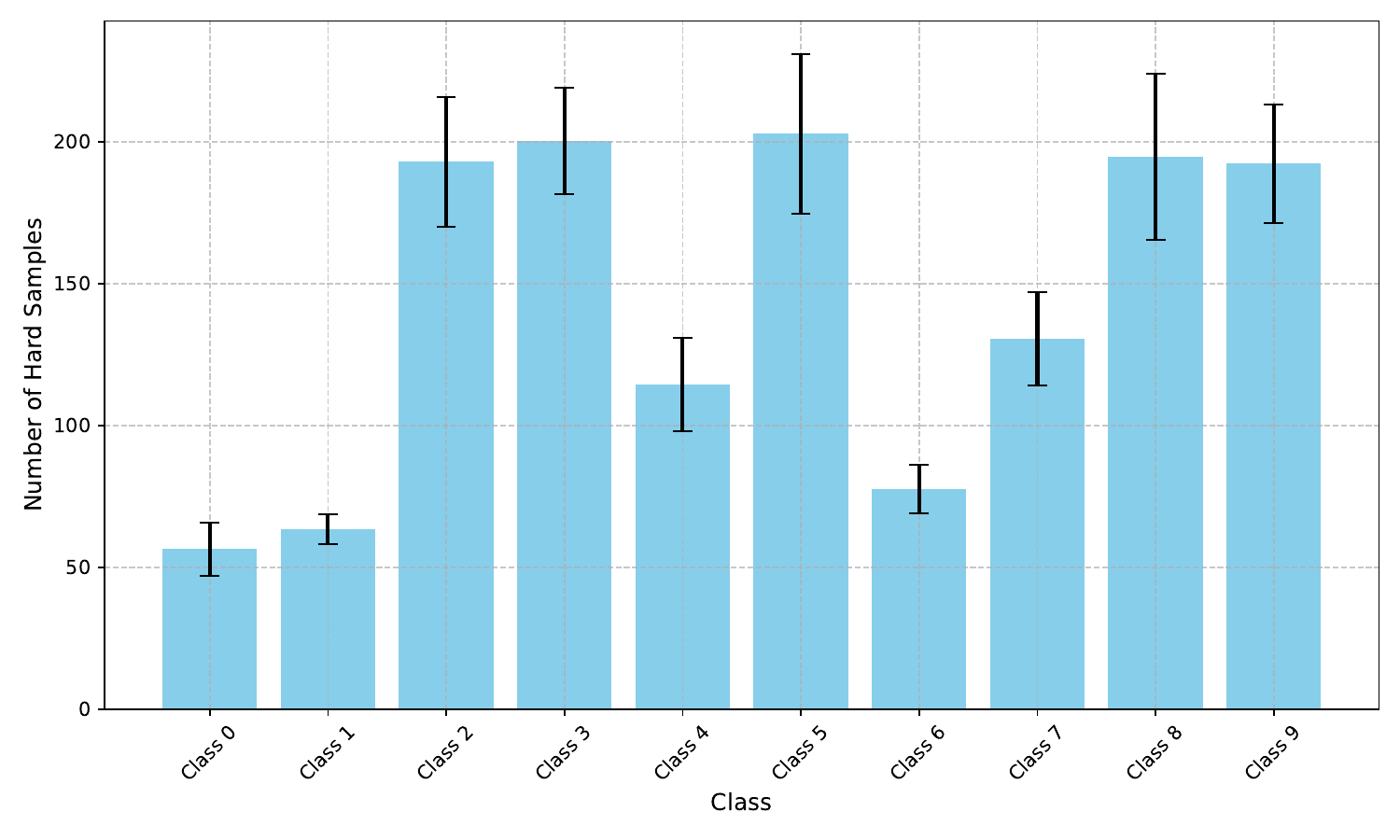}
        \caption{Distribution of hard samples among classes obtained via straggler-based method on MNIST}
        \label{fig:8a}
    \end{subfigure}
    \centering
    \begin{subfigure}{5cm}
        \centering
        \includegraphics[width=5cm]{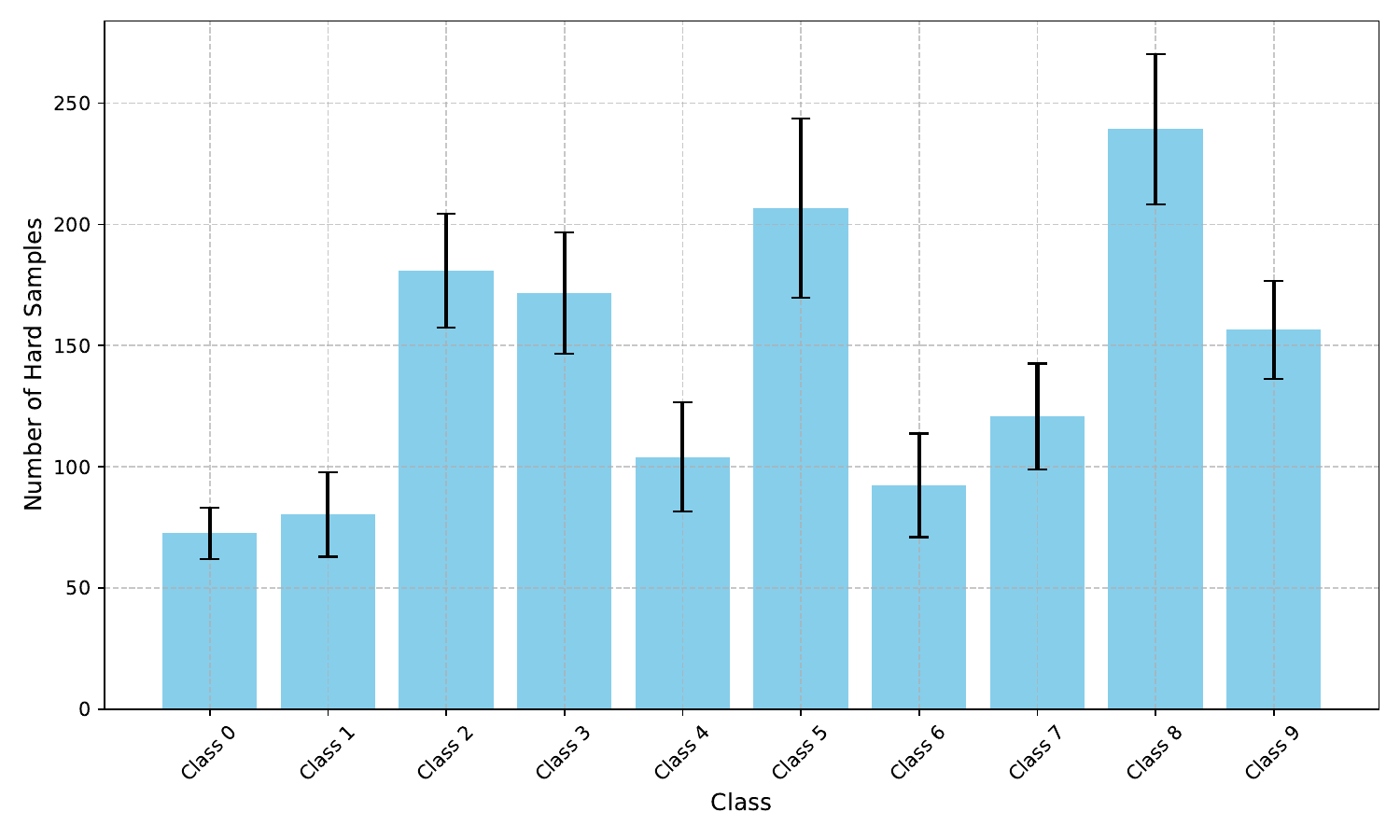}
        \caption{Distribution of hard samples among classes obtained via confidence-based method on MNIST}
        \label{fig:8b}
    \end{subfigure}
    \centering
    \begin{subfigure}{5cm}
        \centering
        \includegraphics[width=5cm]{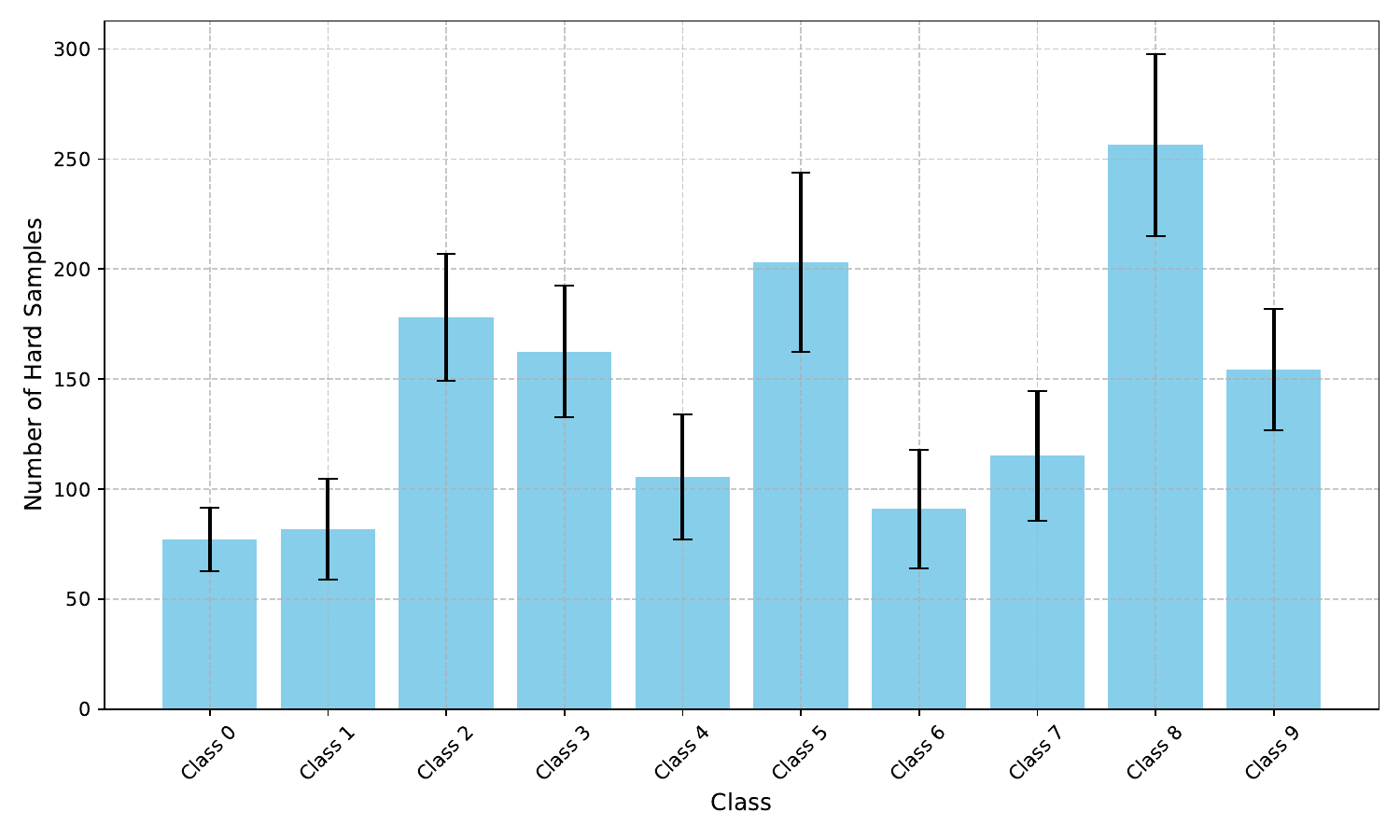}
        \caption{Distribution of hard samples among classes obtained via energy-based method on MNIST}
        \label{fig:8c}
    \end{subfigure}
    \centering
    \begin{subfigure}{5cm}
        \centering
        \includegraphics[width=5cm]{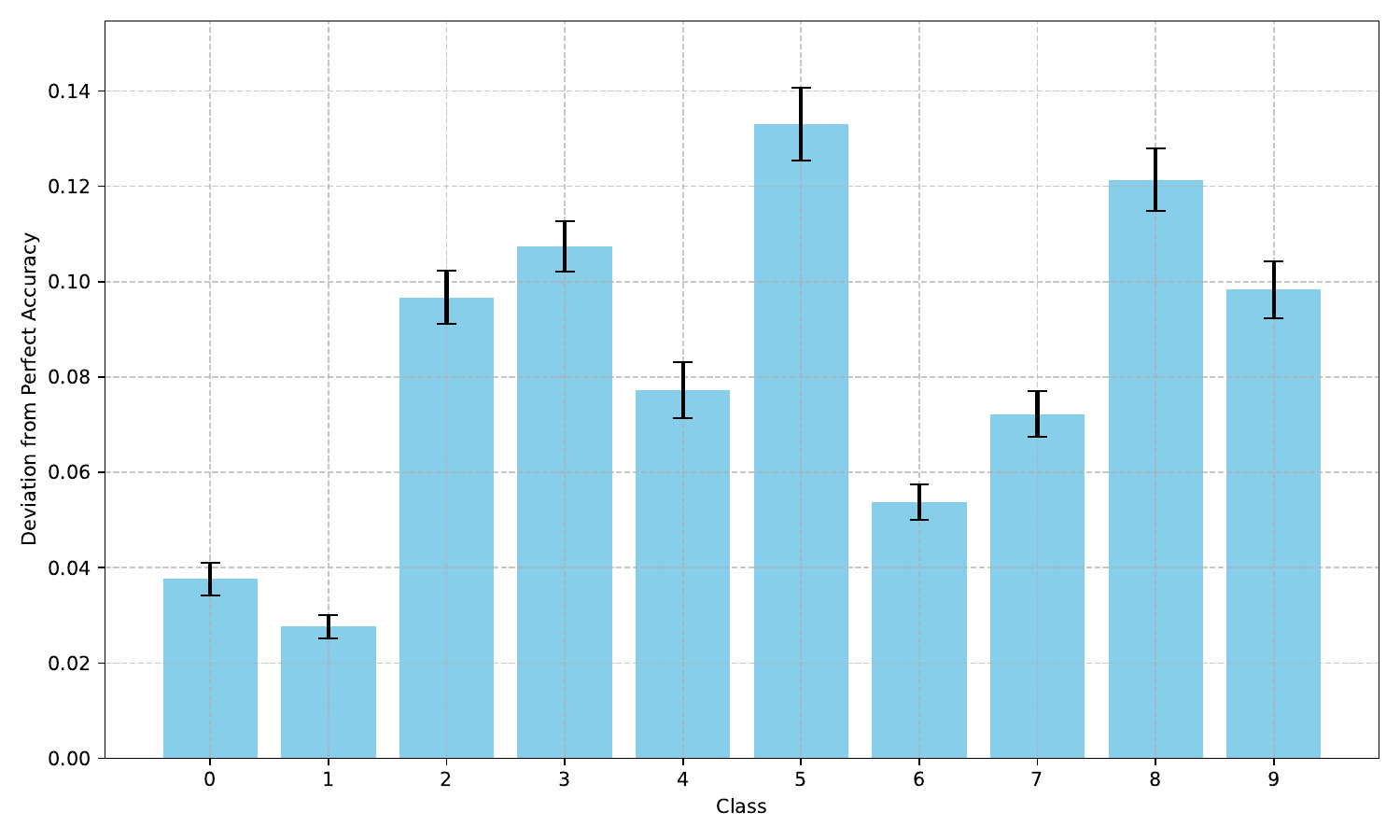}
        \caption{Class-level accuracies on MNIST}
        \vspace{0.9cm}
    \end{subfigure}
    \centering
    \begin{subfigure}{5cm}
        \centering
        \includegraphics[width=5cm]{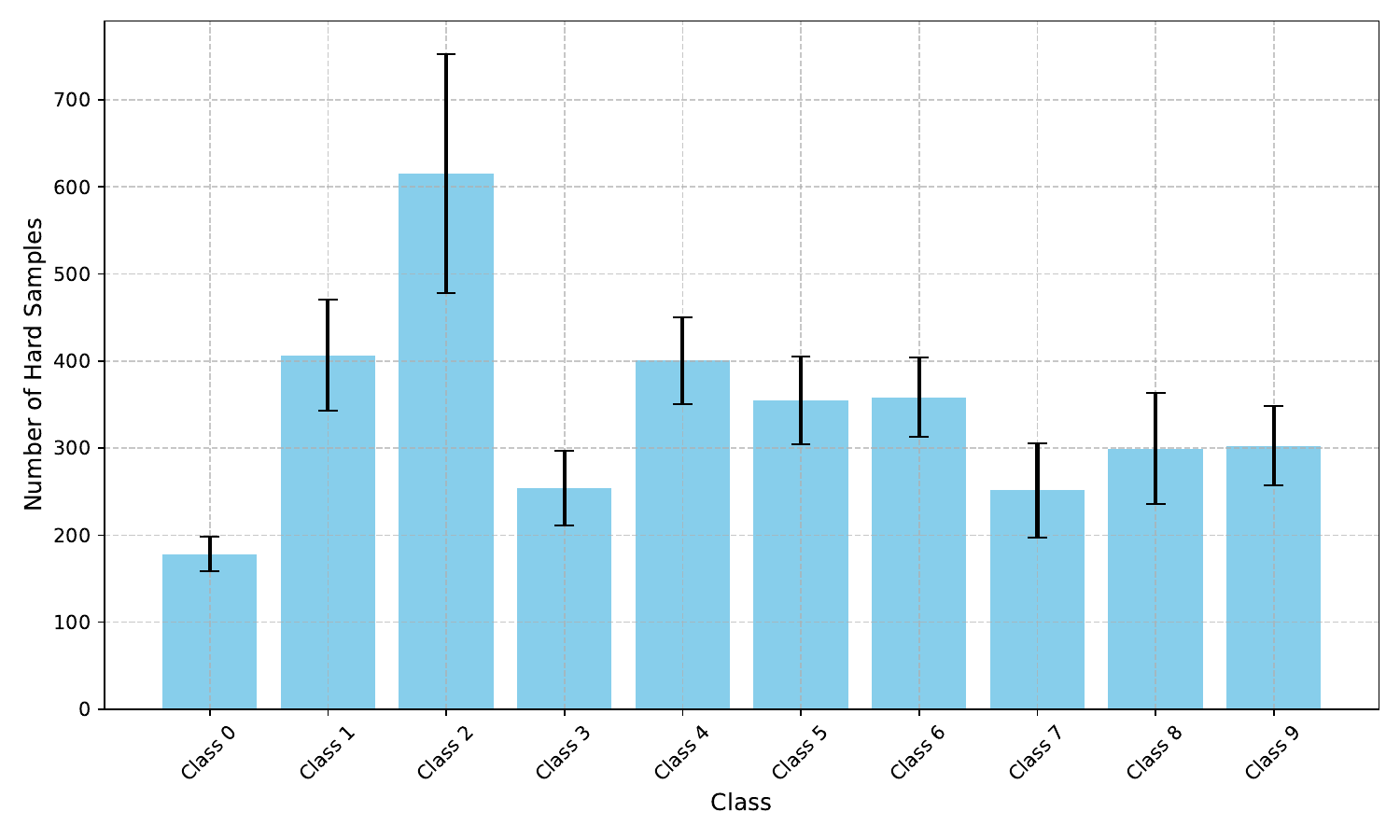}
        \caption{Distribution of hard samples among classes obtained via straggler-based method on KMNIST}
    \end{subfigure}
    \centering
    \begin{subfigure}{5cm}
        \centering
        \includegraphics[width=5cm]{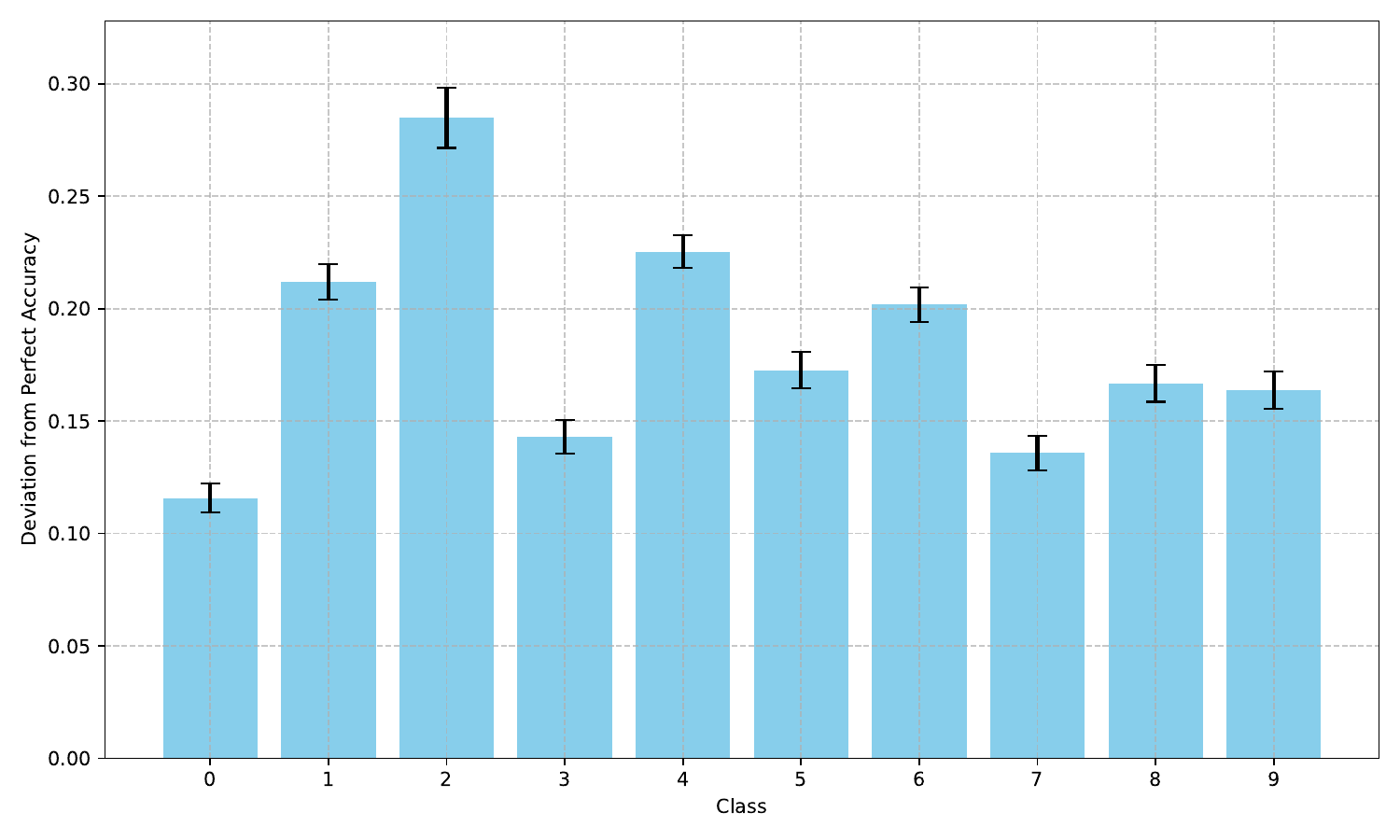}
        \caption{Class-level accuracies on KMNIST}
        \vspace{0.9cm}
    \end{subfigure}
    \centering
    \begin{subfigure}{5cm}
        \centering
        \includegraphics[width=5cm]{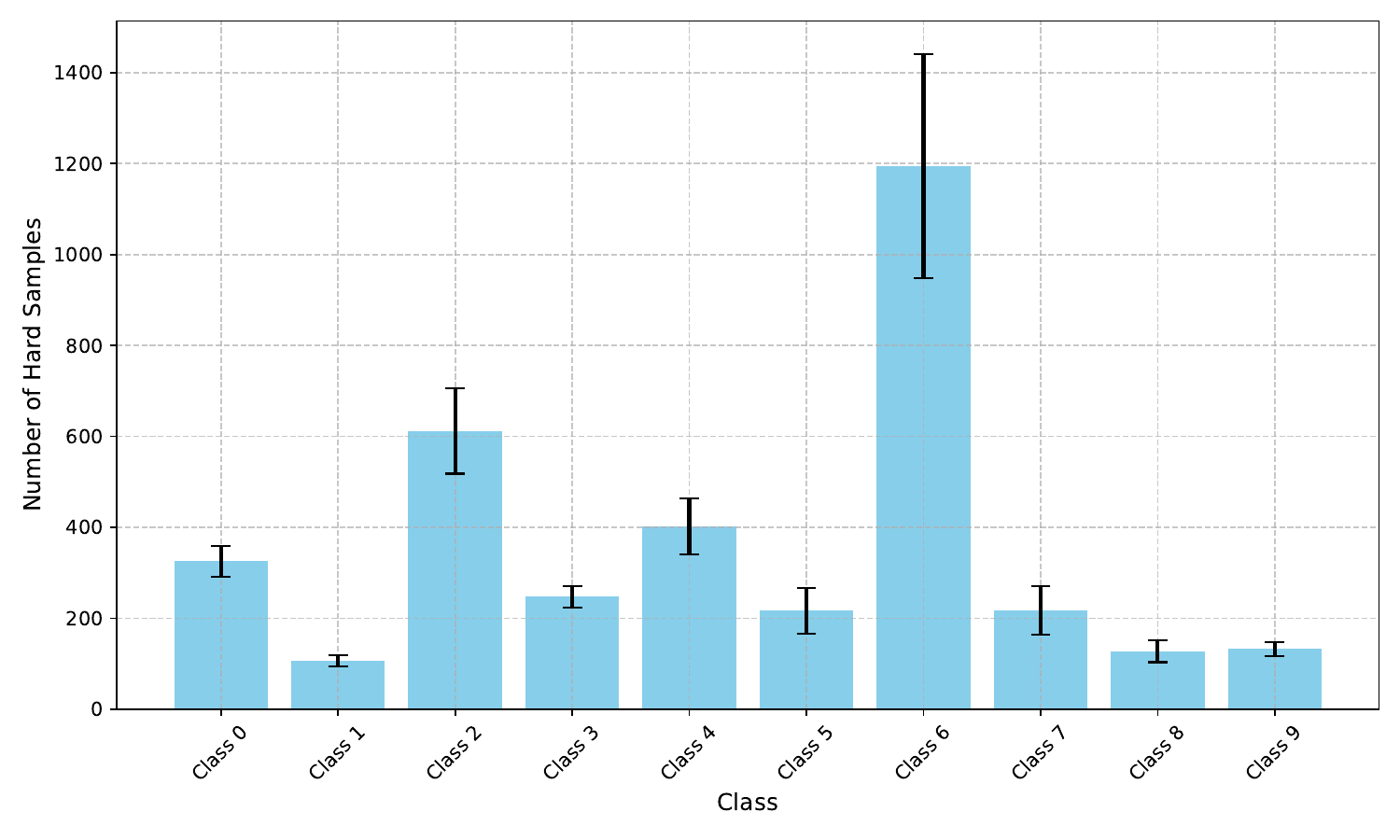}
        \caption{Distribution of hard samples among classes obtained via straggler-based method on FashionMNIST}
    \end{subfigure}
    \centering
    \begin{subfigure}{5cm}
        \centering
        \includegraphics[width=5cm]{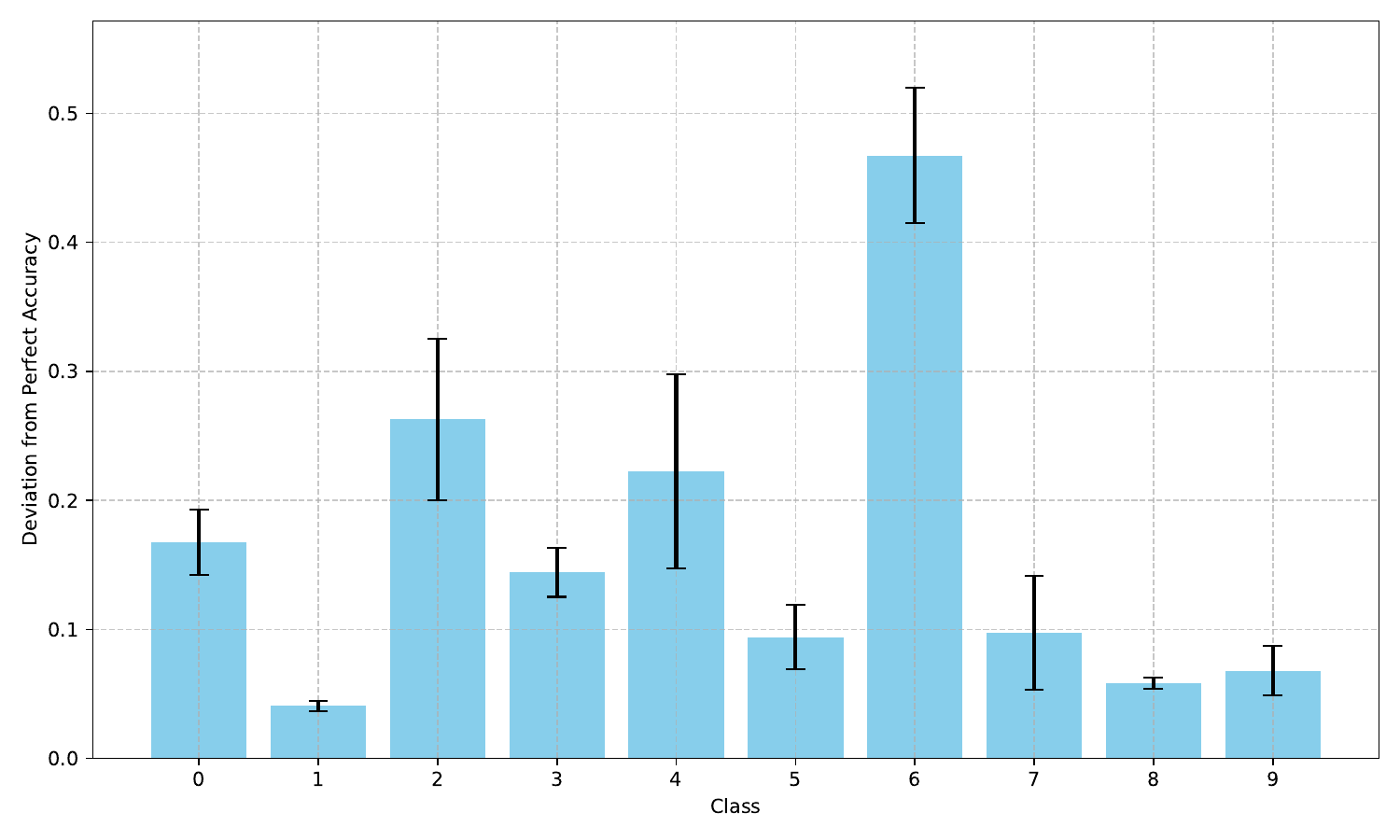}
        \caption{Class-level accuracies on FashionMNIST}
        \vspace{0.9cm}
    \end{subfigure}
    \caption{Examining Figures \ref{fig:fig2} and \ref{fig:fig7} leads us to the natural conclusion that the dynamics of class manifolds vary from one class to another. As a result, we can anticipate that, even though the datasets we have studied are balanced (meaning there is no between-class data imbalance), the distribution of hard samples among the classes is not uniform. This phenomenon occurs regardless of the method used for identifying hard samples (a-c). However, we find that confidence-based (b) and energy-based (c) methods yield a very similar distribution of hard samples, which diverges from that obtained via the straggler-based approach. The close resemblance between methods (b) and (c) arises from an $87.42\% \pm 1.09$ overlap between the hard samples identified by confidence- and energy-based approaches (averaged over $15$ runs). This overlap decreases to $45.67\% \pm 3.07$ for straggler- vs energy-based approaches and $49.48\% \pm 3.04$ for straggler- vs confidence-based approaches. One reason behind the non-uniform distribution of hard samples among classes is that not all classes are equally easy to learn. To demonstrate this, we conducted a simple experiment where we trained a network (using the same hyperparameters as in previous experiments) $100$ times on MNIST and reported the class-level accuracies in (d). We also present the distribution of hard samples obtained with the straggler-based approach, and the distribution of class-level accuracies on KMNIST (e-f) and FashionMNIST (g-h). Performing the Pearson correlation analysis reveals a correlation of over 0.9 between the class-level error rates and class-level distribution of hard samples.}
    \label{fig:final_figure}
\end{figure}

We also measure the distribution of hard samples across the classes. We find that the distribution of hard samples between classes is not uniform (as we show in Figure \ref{fig:final_figure}). We suspect that the main reason behind it is the fact that not all classes are equally easy to learn. Running simple experiments on MNIST, KMNIST and FashionMNIST, in which we train 100 networks (using the same training configuration and hyperparameters as in previous experiments), shows that there exists a very strong correlation between the class-level number of hard samples and class-level error rate. Performing the Pearson correlation analysis backs those visual observations. We find a correlation of 0.9507 for straggler-based methods, 0.9381 for confidence-based methods, and 0.9075 for energy-based methods for the MNIST dataset. For KMNIST and FashionMNIST the correlation is 0.9577 and 0.9856, respectively.

Finally, we also investigate how largely the hard samples identified by different methods overlap. We find that there is an $87.42\% \pm 1.09$ overlap between the hard samples identified by confidence- and energy-based methods. This overlap reduces to $45.67\% \pm 3.07$ when comparing straggler- and energy-based methods, and to $49.48\% \pm 3.04$ between straggler- and confidence-based methods. This explains the variations in observed in-class data imbalance. This also explains why Figures \ref{fig:8b} and \ref{fig:8c} are so closely correlated to each other, and why both of them differ so much from the results in Figure \ref{fig:8a}.

\end{document}